\documentclass[journal]{IEEEtran}
\usepackage{cite}
\ifCLASSINFOpdf
   \usepackage[pdftex]{graphicx}
\else
\fi
\usepackage{amsmath, amssymb}
\ifCLASSOPTIONcompsoc
 \usepackage[caption=false,font=normalsize,labelfont=sf,textfont=sf]{subfig}
\else
 \usepackage[caption=false,font=footnotesize]{subfig}
\fi
\usepackage{algorithm}
\usepackage[noend]{algpseudocode}

\usepackage{bm}
\DeclareMathOperator{\argmax}{arg\,max}
 
\newcommand{\minus}{\scalebox{0.75}[1.0]{$-$}}

\usepackage{booktabs}
\usepackage{hyperref}
\usepackage[stable]{footmisc}
\usepackage{balance}

\begin{document}
\title{Ensemble Quantile Networks: Uncertainty-Aware Reinforcement Learning with Applications in Autonomous Driving}
\author{Carl-Johan~Hoel,~Krister~Wolff,~and~Leo~Laine%
\thanks{
This work was partially supported by the Wallenberg Artificial Intelligence, Autonomous Systems, and Software Program (WASP), funded by the Knut and Alice Wallenberg Foundation}%
\thanks{C. J. Hoel and L. Laine are with Chalmers University of Technology, Gothenburg, Sweden and with Volvo Group, Gothenburg, Sweden (\mbox{e-mail}: \{carl-johan.hoel, leo.laine\}@chalmers.se). }%
\thanks{K. Wolff is with Chalmers University of Technology, Gothenburg, Sweden (\mbox{e-mail}: \mbox{krister.wolff@chalmers.se})}%
}
\maketitle

\begin{abstract}

Reinforcement learning (RL) can be used to create a decision-making agent for autonomous driving. However, previous approaches provide only black-box solutions, which do not offer information on how confident the agent is about its decisions. An estimate of both the aleatoric and epistemic uncertainty of the agent's decisions is fundamental for real-world applications of autonomous driving.
Therefore, this paper introduces the Ensemble Quantile Networks (EQN) method, which combines distributional RL with an ensemble approach, to obtain a complete uncertainty estimate. The distribution over returns is estimated by learning its quantile function implicitly, which gives the aleatoric uncertainty, whereas an ensemble of agents is trained on bootstrapped data to provide a Bayesian estimation of the epistemic uncertainty.
A criterion for classifying which decisions that have an unacceptable uncertainty is also introduced.
The results show that the EQN method can balance risk and time efficiency in different occluded intersection scenarios, by considering the estimated aleatoric uncertainty. 
Furthermore, it is shown that the trained agent can use the epistemic uncertainty information to identify situations that the agent has not been trained for and thereby avoid making unfounded, potentially dangerous, decisions outside of the training distribution.

\end{abstract}


\IEEEpeerreviewmaketitle

\section{Introduction}
\label{sec:introduction}

\IEEEPARstart{A}{decision-making} agent for an autonomous vehicle needs to handle a diverse set of environments and situations, while interacting with other traffic participants and considering uncertainty.
A machine learning approach for creating a general decision-making agent is compelling, since it is not feasible to manually predict all situations that can occur and code a suitable behavior for each and every one of them. However, a drawback of learning-based agents is that they typically provide a black-box solution, which only outputs a decision for a given situation. It would be desirable if the agent also could provide an estimate of its confidence level, or equivalently, the estimated uncertainty of its decisions.

Uncertainty can be divided into two categories: aleatoric and epistemic uncertainty~\cite{Kiureghian2009, Hullermeier2021}, where both are important to consider in many decision-making problems.
Aleatoric uncertainty refers to the inherent randomness of an outcome and can therefore not be reduced by observing more data. For example, when approaching an occluded intersection, there is an aleatoric uncertainty in if, or when, another vehicle will enter the intersection.
To estimate the aleatoric uncertainty is important, since such information can be used to make risk-aware decisions.
Contrarily, epistemic uncertainty arises due to a lack of knowledge and can be reduced by observing more data. For example, epistemic uncertainty appears if a decision-making agent has been trained to only handle `normal' driving situations and then faces a speeding driver or an accident. 
An estimate of the epistemic uncertainty provides insight into which situations the trained agent does not know how to handle and can be used to increase the safety~\cite{McAllister2017}. The epistemic uncertainty estimate could also be used to concentrate the training process to situations where the agent needs more training~\cite{Osband2016}.

Reinforcement learning (RL) provides a learning-based approach to create decision-making agents, which could potentially scale to all driving situations.
Many recent studies have applied RL to autonomous driving, for example, by using the Deep Q-Network (DQN) algorithm in intersections and highway situations~\cite{Tram2018, Isele2018, Hoel2018}, by using a policy gradient method for lane merging~\cite{Shalev2016}, or combining RL with Monte Carlo tree search~\cite{Hoel2019}.
A majority of these studies perform both the training and evaluation in simulated environments, whereas some train the agent in simulations and then apply the trained agent in the real world~\cite{Pan2017, Bansal2018}, or for some limited cases, the training itself is also performed in the real world~\cite{Kendall2019}. 
Overviews of RL for autonomous driving are given by Kiran et al.~\cite{Kiran2021} and by Zhu et al.~\cite{Zhu2021}.
However, previous studies do not estimate the aleatoric or the epistemic uncertainty of the decision that the trained agent recommends. One exception is the study by Bernhard et al., where a distributional RL approach is used to create a risk-sensitive decision-making agent~\cite{Bernhard2019}. However, the method is not applied in a theoretically consistent way and can therefore cause arbitrary decisions, which is further discussed in Sect.~\ref{sec:discussion} of this paper.

Bayesian probability theory can be used to estimate the epistemic uncertainty~\cite{Kochenderfer2015}. In the autonomous driving field, Bayesian deep learning has been used for, e.g.,  image segmentation~\cite{Kendall2017} and end-to-end learning~\cite{Michelmore2018}. For RL, Bayesian techniques have been used to balance the exploration vs. exploitation trade-off~\cite{Dearden1998}, and more recent work has addressed similar problems in deep RL~\cite{Osband2018}. 
Furthermore, the aleatoric uncertainty of a decision can be obtained through distributional RL, which aims to model the distribution over returns, instead of only the mean return, as in standard RL~\cite{Sobel1982},~\cite{Morimura2010}. For example, Bellemare et al. introduced a method for estimating the probability of a discrete set of returns~\cite{Bellemare2017}, which also has been further developed for continuous control tasks~\cite{Barth-Maron2018}.

In contrast to the related work, this paper presents methods for training an RL agent for autonomous driving, in which the trained agent provides an estimate of the epistemic and the aleatoric uncertainty of its decisions. The epistemic uncertainty estimate is obtained through a Bayesian RL approach, which extends and further analyses the approach from two studies by the authors of this paper~\cite{Hoel2020_1},~\cite{Hoel2020_2}. This method, based on the work by Osband et al.~\cite{Osband2018}, uses an ensemble of neural networks with additive random prior functions to obtain a posterior distribution over the expected return (Sect.~\ref{sec:epistemic}). The aleatoric uncertainty is obtained through a distributional RL approach, based on the work by Dabney et al.~\cite{Dabney2018}, which estimates the probability distribution over returns by implicitly learning its quantile function\footnote{The quantile function is the inverse of the cumulative distribution function for a continuous random variable.}. This method also allows the agent to be trained in a risk-aware manner (Sect.~\ref{sec:aleatoric}).
Furthermore, this paper introduces the Ensemble Quantile Networks (EQN) method, which combines the two previously mentioned approaches, in order to provide a complete uncertainty estimate of both the aleatoric and epistemic uncertainty of an agent's decisions (Sect.~\ref{sec:combination}).
The performance of the proposed methods is tested and analyzed in different intersection scenarios (Sect.~\ref{sec:implementation}), where the results show that while they outperform the standard DQN method, the epistemic uncertainty estimate can be used to choose less risky actions in unknown situations, and the distributional risk-aware approach allows a trade-off between risk and time efficiency (Sect.~\ref{sec:results}). Another potential use for the epistemic uncertainty information is to identify situations that should be added to the training process. Further properties of the proposed approaches are discussed in Sect.~\ref{sec:discussion}. 
The code that was used to implement the different algorithms and the simulated scenarios is available on GitHub~\cite{sourceCode2021}.

The main contributions of this paper are:
\begin{enumerate}
    \item Methods for estimating either the aleatoric or the epistemic uncertainty of a trained agent, together with confidence criteria, which can be used to identify situations with high uncertainty (Sect.~\ref{sec:aleatoric_criterion}, ~\ref{sec:epistemic_criterion}).
    \item The introduction of the EQN algorithm, which simultaneously quantifies both the aleatoric and the epistemic uncertainty of a trained agent (Sect.~\ref{sec:combination}).
    \item A detailed description of how the proposed methods can be applied to an autonomous driving setting (Sect.~\ref{sec:implementation}).
    \item A qualitative and quantitative performance analysis of the proposed methods for different intersection scenarios (Sect.~\ref{sec:results}).
\end{enumerate}

\section{Approach}
\label{sec:approach}

This section first gives a brief introduction to RL and its notation, followed by a description of how an aleatoric and epistemic uncertainty estimate can be obtained. The details on how these approaches can be applied to driving in an intersection scenario follows in Sect.~\ref{sec:implementation}.

\subsection{Reinforcement learning}
Reinforcement learning is a branch of machine learning, where an agent learns a policy $\pi(s)$ from interacting with an environment~\cite{Sutton2018}. The policy describes which action $a$ to take in state $s$. The environment then transitions to a new state $s'$ and the agent receives a reward $r$. 
The decision-making problem that the RL agent tries to solve is often modeled as a Markov decision process (MDP), defined by the tuple $(\mathcal{S}, \mathcal{A}, R, T, \gamma)$, where $\mathcal{S}$ is the state space, $\mathcal{A}$ is the action space, $R$ is a reward model, $T$ is the state transition model, and $\gamma$ is a discount factor. The goal of the agent is to maximize the expected future discounted return $\mathbb{E}[R_t]$, for every time step $t$, where
\begin{align}
    R_t = \sum_{k=0}^\infty \gamma^k r_{t+k}.
\end{align}

The value of taking action $a$ in state $s$ and then following policy $\pi$ is defined by the state-action value function
\begin{align}
    Q^\pi(s,a) = \mathbb{E}[R_t|s_t=s, a_t=a, \pi],
\end{align}
where the $Q$-values for the optimal policy $\pi^*$ are defined as $Q^*(s,a)=\max_\pi Q^\pi(s,a)$.
The DQN algorithm aims to approximate the optimal state-action value function $Q^*$ by a neural network with weights $\theta$, such that $Q(s,a;\theta) \approx Q^*(s,a)$ \cite{Mnih2015}.
Based on the Bellman equation, the temporal difference (TD) error
\begin{align}
\label{eq:td_dqn}
    \delta_t = r_t + \gamma \max_{a} Q(s_{t+1},a;\theta^-)
    - Q(s_t,a_t;\theta)
\end{align}
is used to optimize the weights by iteratively minimizing the loss function $L_\mathrm{DQN}(\theta) = \mathbb{E}_M [\delta_t^2]$. The loss is calculated for a mini-batch $M$ of experiences, where each experience consists of the tuple $(s_t, a_t, r_t, s_{t+1})$, and the network weights $\theta$ are updated by stochastic gradient descent (SGD). Finally, $\theta^-$ is a target network that is updated regularly.

\subsection{Aleatoric uncertainty estimation}
\label{sec:aleatoric}

In contrast to $Q$-learning, distributional RL aims to learn not only the expected return, but the distribution over returns~\cite{Bellemare2017}. This distribution is represented by the random variable
$Z^\pi(s,a) = R_t$, given $s_t=s$, $a_t=a$, and policy $\pi$,
where the mean is the traditional value function, i.e., $Q^\pi(s,a) = \mathbb{E}[Z^\pi(s,a)]$.
The distribution over returns represents the aleatoric uncertainty of the outcome, which can be used to estimate the risk in different situations and to train an agent in a risk-sensitive manner.

The implicit quantile networks (IQN) approach~\cite{Dabney2018} to distributional RL uses a neural network to implicitly represent the quantile function $F_Z^{-1}(\tau)$ of the random variable $Z$ and then update the weights of the network with quantile regression. For ease of notation, define $Z_\tau := F_Z^{-1}(\tau)$, and note that for $\tau \sim \mathcal{U}(0,1)$ the sample $Z_\tau(s,a) \sim Z(s,a)$. The TD-error for two quantile samples, $\tau, \tau' \sim \mathcal{U}(0,1)$, is
\begin{align}
    \delta_t^{\tau, \tau'} = r_t + \gamma Z_{\tau'}{\left(s_{t+1},\pi^*(s_{t+1});\theta^-\right)} - Z_\tau(s_t,a_t;\theta),
\end{align}
where $\pi^*(s) = \argmax_a{Q(s,a)}$.
A sample-based estimate of $\pi^*(s)$ is obtained from $K_{\tau}$ samples of $\tilde{\tau} \sim \mathcal{U}(0,1)$, as
\begin{align}
\label{eq:pibeta}
    \tilde{\pi}(s) = \argmax_a{\frac{1}{K_{\tau}} \sum_{k=1}^{K_{\tau}} Z_{\tilde{\tau}_k}(s,a;\theta)}.
\end{align}
For a pair of quantiles $\tau, \tau'$, the quantile Huber regression loss~\cite{Dabney_QR}, with threshold $\kappa$, is calculated as
\begin{align}
    \rho_\kappa(\delta_t^{\tau,\tau'}) & = |\tau - \mathbb{I}\{\delta_t^{\tau,\tau'}<0\}|\frac{\mathcal{L}_\kappa(\delta_t^{\tau,\tau'})}{\kappa}.
\end{align}
Here, $\mathcal{L}_\kappa(\delta_t^{\tau,\tau'})$ is the Huber loss~\cite{Huber1964}, defined as
\begin{align}
    \mathcal{L}_\kappa(\delta_t^{\tau,\tau'}) = &\begin{cases} 
    \frac{1}{2}{(\delta_t^{\tau,\tau'})}^2, & \mathrm{if\ } |\delta_t^{\tau,\tau'}| \leq \kappa, \\
    \kappa(|\delta_t^{\tau,\tau'}|-\frac{1}{2}\kappa), & \mathrm{otherwise},
    \end{cases}
\end{align}
which gives a smooth gradient as $\delta_t^{\tau,\tau'} \to 0$.
The full loss function $L_\mathrm{IQN}(\theta)$ is obtained from a mini-batch $M$ of sampled experiences, in which the quantiles $\tau$ and $\tau'$ are sampled $N$ and $N'$ times, respectively, according to
\begin{align}
    L_\mathrm{IQN}(\theta) = \mathbb{E}_M{\left[\frac{1}{N'}\sum_{i=1}^N \sum_{j=1}^{N'} \rho_\kappa{\left(\delta_t^{\tau_i, \tau_j'}\right)}\right]}.
\end{align}
The full training process of the IQN method is outlined in Algorithm~\ref{alg:iqn_training}.

\begin{algorithm}[!t]
    \caption{IQN training process}\label{alg:iqn_training}
    \begin{algorithmic}[1]
        \State Initialize $\theta$ randomly
        \State $m \gets \{\}$
        \State $t \gets 0$
        \While{network not converged}
            \State $s_t \gets $ initial random state
            \While{episode not finished}
                \If{$e \sim \mathcal{U}(0,1) < \epsilon$}
                    \State $a_t \gets \mathrm{random\ action}$
                \Else
                    \State $\tau_1, \ldots, \tau_{K_{\tau}} \overset{\mathrm{i.i.d.}}\sim \mathcal{U}(0,\alpha)$
                    \State $a_t \gets \argmax_{a} \frac{1}{K_{\tau}} \sum_{k=1}^{K_{\tau}} Z_{\tau_k} (s_t,a)$
                \EndIf
                \State $s_{t+1}, r_t \gets $ \Call{StepEnvironment}{$s_t, a_t$}
                   \State $m \gets m \cup \{(s_t, a_t, r_t, s_{t+1})\}$
                   \State $M \gets $ sample from $m$
                   \State update $\theta$ with SGD and loss $L_\mathrm{IQN}(\theta)$ 
                \State $t \gets t + 1$
            \EndWhile
        \EndWhile
    \end{algorithmic}
\end{algorithm}

\begin{figure}[!t]
    \centering
    \subfloat[Probability density function.]{\includegraphics[width=0.48\columnwidth]{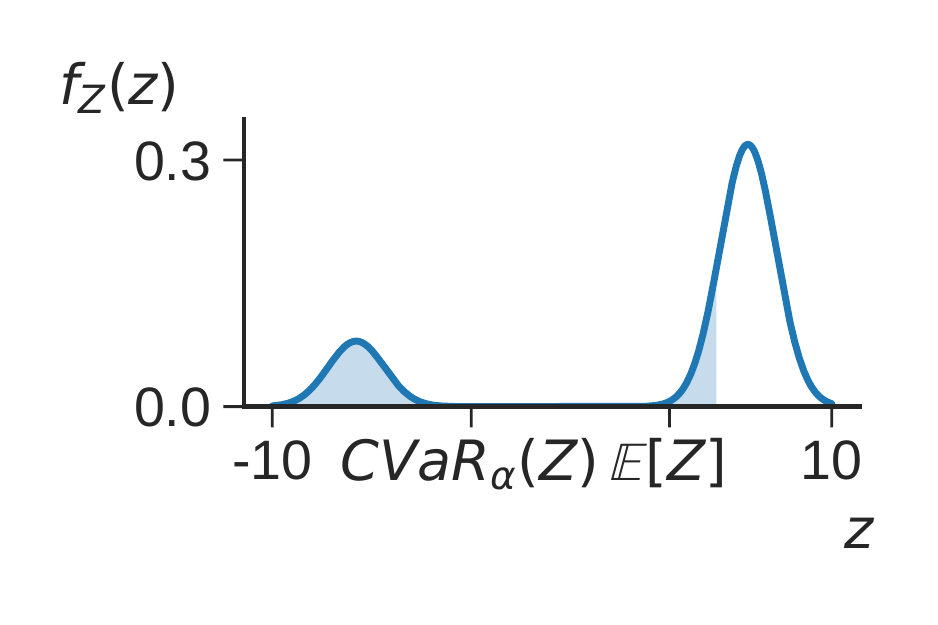}}
    \hfil
    \subfloat[Cumulative distribution function.]{\includegraphics[width=0.48\columnwidth]{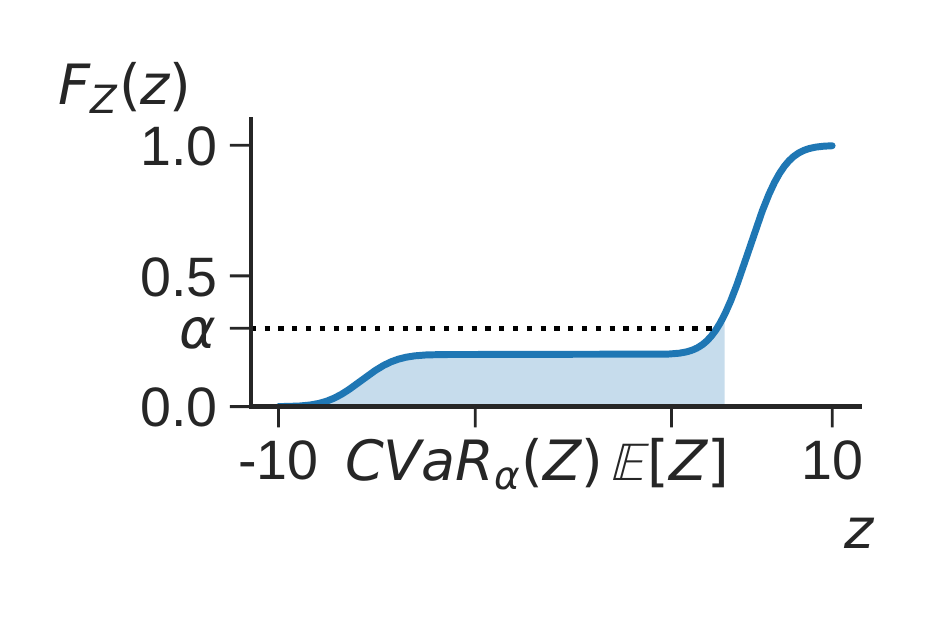}}
    \hfil
    \subfloat[Quantile function.]{\includegraphics[width=0.98\columnwidth]{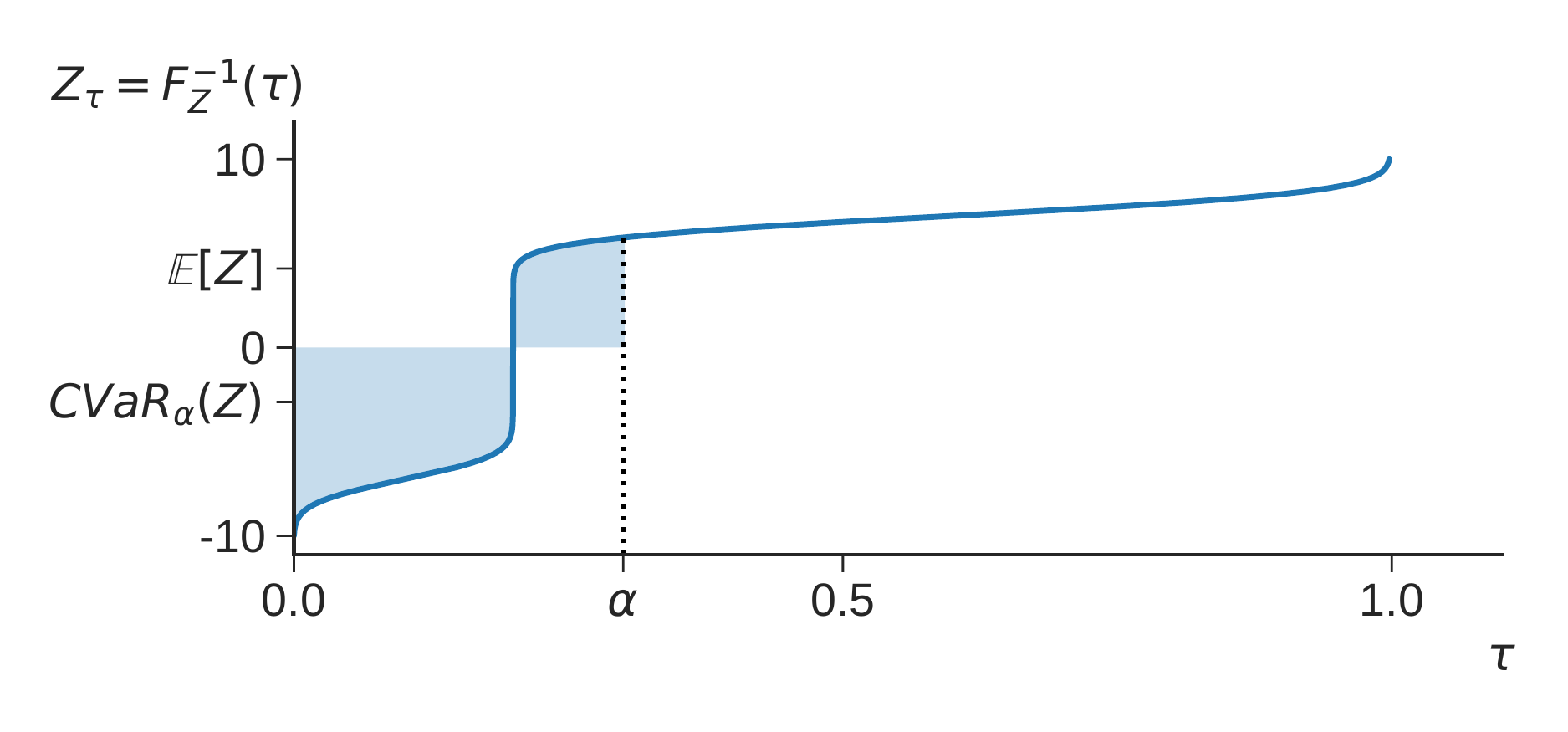}}
    \caption{Illustration of the $\mathrm{CVaR}_\alpha$ risk measure. The shaded regions represent quantiles $\tau \in [0, \alpha]$, here for $\alpha=0.3$.}
    \label{fig:cvar}
\end{figure}

\subsubsection{Risk-sensitive RL}
\label{sec:risk-sensitiveRL}

In the present context, risk refers to the aleatoric uncertainty of the potential outcome of an action. 
Eq.~\ref{eq:pibeta} represents a risk-neutral policy, which maximizes the $Q$-values.
An alternative risk-averse policy is obtained by instead choosing the action that maximizes the conditional value-at-risk (CVaR)~\cite{Rockafellar2002}, where
\begin{align}
    \mathrm{CVaR}_\alpha{\left(Z(s,a)\right)} = \mathbb{E}_{\tilde{\tau} \sim U([0,\alpha])} {\left[Z_{\tilde{\tau}}(s,a)\right]}.
\end{align}
The CVaR approach selects actions that maximize the mean outcome of quantiles less than $\alpha$, which is graphically illustrated in Fig.~\ref{fig:cvar}.
A detailed description of the CVaR approach and its use in solving MDPs is presented by Chow et al.~\cite{Chow2014}.
Majumdar et al. further discuss the use of different distortion risk measures in robotics~\cite{Majumdar2017}.

\subsubsection{Uncertainty criterion}
\label{sec:aleatoric_criterion}

Dabney et al. show that the IQN method can achieve state-of-the-art results on the \mbox{Atari-57} benchmark and reason about the performance of risk-sensitive training for a few of the Atari games~\cite{Dabney2018}. However, as introduced in this paper, the estimated distribution over returns of the fully trained IQN agent can also be used to quantify the aleatoric uncertainty of a decision. One such uncertainty measure is the variance of the estimated returns for the evenly distributed sample set $\tau_{\sigma} = \{i/K_\tau \mid i\in[1,K_\tau]\}$. A threshold $\sigma_\mathrm{a}^2$ can then be defined, such that the agent classifies a decision with a higher variance in returns as uncertain.
In this study, the benefit of the introduced uncertainty classification is demonstrated by choosing a predefined backup policy $\pi_\mathrm{backup}(s)$ if the sample variance is higher than the threshold, i.e., the fully trained agent follows the policy
\begin{align}
    \pi_{\sigma_\mathrm{a}}(s) \smash{=} 
    &\begin{cases} \argmax_a{\mathbb{E}_{\tau_\sigma}[Z_\tau(s,a)]}\rlap{,} & \mathrm{if\ } \mathrm{Var}_{\tau_\sigma}[Z_\tau(s,a)] \smash{<} \sigma^2_\mathrm{a},\\
    \pi_\mathrm{backup}(s), & \mathrm{otherwise}.
    \end{cases}
\end{align}

\subsection{Epistemic uncertainty estimation}
\label{sec:epistemic}

The DQN algorithm gives a maximum likelihood estimate of the $Q$-values, and the IQN algorithm outputs a maximum likelihood estimate of the distribution over returns. However, neither of these algorithms considers the epistemic uncertainty of the recommended actions.
Statistical bootstrapping~\cite{Efron1982} can be used to train an ensemble of neural networks on different subsets of the available data, which provides a distribution over the estimated $Q$-values~\cite{Osband2016}. A better Bayesian posterior can be obtained by adding a randomized prior function (RPF) to each ensemble member, which creates a larger output diversity outside of the training distribution~\cite{Osband2018}. The $Q$-values of ensemble member $k$ is then given by the sum of two neural networks, $f$ and $p$, with identical architecture, i.e.,
\begin{align}
    Q_k(s,a) = f(s,a;\theta_k) + \beta p(s,a;\hat{\theta}_k).
\end{align}
The parameters $\theta_k$ are trained, whereas, importantly, the parameters of the prior function $\hat{\theta}_k$ are fixed during the training process. A hyperparameter $\beta$ scales the relative importance of the networks. The additional prior network of the RPF method gives a slightly modified TD-error compared to the DQN method (Eq.~\ref{eq:td_dqn}), which results in the loss function
\begin{align}
    \label{eq:loss_rpf}
    L_\mathrm{RPF}(\theta_k) = \mathbb{E}_M \Big[ & (r_t + \gamma \max_{a} (f_{\theta^-_k}+\beta p_{\hat{\theta}_k})(s_{t+1},a) \nonumber \\
    & - (f_{\theta_k}+ \beta p_{\hat{\theta}_k})(s_t,a_t) )^2 \Big].
\end{align}

\begin{algorithm}[!t]
    \caption{Ensemble RPF training process}\label{alg:rpf_training}
    \begin{algorithmic}[1]
        \For{$k \gets 1$ to $K$}
            \State Initialize $\theta_k$ and $\hat{\theta}_k$ randomly
            \State $m_k \gets \{\}$
        \EndFor
        \State $t \gets 0$
        \While{networks not converged}
            \State $s_t \gets $ initial random state
            \State $\nu \sim \mathcal{U}\{1,K\}$
            \While{episode not finished}
                \State $a_t \gets \argmax_{a} Q_\nu(s_t,a)$
                \State $s_{t+1}, r_t \gets $ \Call{StepEnvironment}{$s_t, a_t$}
                \For{$k \gets 1$ to $K$}
                   \If{$p \sim \mathcal{U}(0,1) < p_\mathrm{add}$}
                      \State $m_k \gets m_k \cup \{(s_t, a_t, r_t, s_{t+1})\}$
                   \EndIf
                   \State $M \gets $ sample from $m_k$
                   \State update $\theta_k$ with SGD and loss $L_\mathrm{RPF}(\theta_k)$
                \EndFor
                \State $t \gets t + 1$
            \EndWhile
        \EndWhile
    \end{algorithmic}
\end{algorithm}

Algorithm~\ref{alg:rpf_training} outlines the training process of the ensemble RPF method. An ensemble of $K$ prior and trainable networks is first initialized randomly. Each ensemble member is also assigned an individual experience replay buffers $m_k$ (although in a practical implementation, the replay buffers can be constructed such that they use negligible more memory than a single shared buffer). For each new training episode, an ensemble member $\nu$ is chosen uniformly at random and is then used to greedily select the actions with the highest $Q$-values throughout the episode. This procedure, which corresponds to an approximate Thompson sampling of the actions, efficiently balances the exploration vs. exploitation trade-off. Each new experience, $(s_t, a_t, r_t, s_{t+1})$, is added to the individual replay buffers $m_k$ with probability $p_\mathrm{add}$. The trainable parameters $\theta_k$ of each ensemble member are then updated through SGD, using a mini-batch $M$ of experiences from the corresponding replay buffer and the loss function in Eq.~\ref{eq:loss_rpf}. 
Finally, when the training process is finished and the agent is tested, the trained agent applies a policy which maximizes the mean $Q$-value of all the ensemble members.

\subsubsection{Uncertainty criterion}
\label{sec:epistemic_criterion}

Osband et al. illustrate the efficient exploration properties of the ensemble RPF algorithm~\cite{Osband2018}, but do not use the estimated distribution over $Q$-values further.
In a similar approach as for the aleatoric uncertainty (Sect.~\ref{sec:aleatoric_criterion}), the variance of the estimated $Q$-values of the ensemble RPF agent can be used to quantify the epistemic uncertainty of a decision, which we introduced in a recent paper~\cite{Hoel2020_1}. 
With this approach, decisions that has a higher variance than a predefined threshold $\sigma_\mathrm{e}^2$ are classified as uncertain. The benefit of the epistemic uncertainty classification is here demonstrated by choosing a predefined backup policy $\pi_\mathrm{backup}(s)$ if the sample variance is higher than the threshold, which means that a trained agent follows the policy
\begin{align}
    \pi_{\sigma_\mathrm{e}}(s) = 
    &\begin{cases} \argmax_{a} \mathbb{E}_k[Q_k(s,a)], & \mathrm{if\ } \mathrm{Var}_k[Q_k(s,a)] < \sigma^2_\mathrm{e}, \\
    \pi_\mathrm{backup}(s), & \mathrm{otherwise}.
    \end{cases}
\end{align}
Further applications of an epistemic uncertainty classification are discussed in Sect.~\ref{sec:discussion}.

\subsection{Aleatoric and epistemic uncertainty estimation}
\label{sec:combination}

A complete uncertainty estimation of both the aleatoric and the epistemic uncertainty can be obtained by combining the properties of the IQN and ensemble RPF methods into a new algorithm, which we call the Ensemble Quantile Networks method. An agent that is trained by the EQN method can then take actions that consider both the inherent uncertainty of the outcome and the model uncertainty in each situation.

As the name suggests, the EQN method uses an ensemble of networks, where each ensemble member $k$ individually estimates the distribution over returns as
\begin{align}
    Z_{k, \tau}(s,a) = f_\tau(s,a;\theta_k) + \beta p_\tau(s,a;\hat{\theta}_k).
\end{align}
Similarly as for the RPF method, $f_\tau$ and $p_\tau$ are neural networks with identical architecture, $\theta_k$ are trainable network parameters, whereas the parameters $\hat{\theta}_k$ are fixed. The TD-error of ensemble member $k$ and two quantile samples, $\tau, \tau' \sim \mathcal{U}(0,1)$, is
\begin{align}
    \delta_{k,t}^{\tau, \tau'} = r_t + \gamma Z_{k,\tau'}{\left(s_{t+1},\tilde{\pi}_k(s_{t+1})\right)} - Z_{k,\tau}(s_t,a_t),
\end{align}
where $\tilde{\pi}_k(s) =  \argmax_a{\frac{1}{K_{\tau}} \sum_{j=1}^{K_{\tau}} Z_{k,\tilde{\tau}_j}(s,a)}$ is a sample-based estimate of the optimal policy.
Quantile Huber regression is applied to a mini-batch of experiences, which gives the loss function
\begin{align}
    L_\mathrm{EQN}(\theta_k) = \mathbb{E}_M{\left[\frac{1}{N'}\sum_{i=1}^N \sum_{j=1}^{N'} \rho_{\kappa}{\left(\delta_{k,t}^{\tau_i, \tau_j'}\right)}\right]}.
\end{align}
For each new training episode, the agent follows the policy $\tilde{\pi}_\nu(s)$ of a randomly selected ensemble member $\nu$.
The full training process of the EQN agent is outlined in Algorithm~\ref{alg:eqn_training}.

\begin{algorithm}[!t]
    \caption{EQN training process}\label{alg:eqn_training}
    \begin{algorithmic}[1]
        \For{$k \gets 1$ to $K$}
            \State Initialize $\theta_k$ and $\hat{\theta}_k$ randomly
            \State $m_k \gets \{\}$
        \EndFor
        \State $t \gets 0$
        \While{networks not converged}
            \State $s_t \gets $ initial random state
            \State $\nu \sim \mathcal{U}\{1,K\}$
            \While{episode not finished}
                \State $\tau_1, \ldots, \tau_{K_{\tau}} \overset{\mathrm{i.i.d.}}\sim \mathcal{U}(0,\alpha)$
                    \State $a_t \gets \argmax_{a} \frac{1}{K_{\tau}} \sum_{k=1}^{K_{\tau}} Z_{\nu, \tau_k} (s_t,a)$
                
                \State $s_{t+1}, r_t \gets $ \Call{StepEnvironment}{$s_t, a_t$}
                \For{$k \gets 1$ to $K$}
                   \If{$p \sim \mathcal{U}(0,1) < p_\mathrm{add}$}
                      \State $m_k \gets m_k \cup \{(s_t, a_t, r_t, s_{t+1})\}$
                   \EndIf
                   \State $M \gets $ sample from $m_k$
                   \State update $\theta_k$ with SGD and loss $L_\mathrm{EQN}(\theta_k)$
                \EndFor
                \State $t \gets t + 1$
            \EndWhile
        \EndWhile
    \end{algorithmic}
\end{algorithm}

\subsubsection{Uncertainty criterion}
\label{sec:combined_criterion}

The EQN agent provides an estimate of both the aleatoric and epistemic uncertainty, based on the variance of the returns and the variance of the \mbox{$Q$-values}. The agent is considered confident about a decision if 
$\mathrm{Var}_{\tau_\sigma}[ \mathbb{E}_k [ Z_{k,\tau}(s,a)]] < \sigma^2_\mathrm{a}$ and $\mathrm{Var}_k[\mathbb{E}_{\tau_\sigma}[Z_{k,\tau}(s,a)]] < \sigma^2_\mathrm{e}$.
The trained agent then follows the policy
\begin{align}
    \pi_{\sigma_\mathrm{a}, \sigma_\mathrm{e}}(s) = 
    &\begin{cases} \argmax_a{ \mathbb{E}_k [ \mathbb{E}_{\tau_\sigma}[Z_{k,\tau}(s,a)]}],
    & \mathrm{if\ confident},\\
    \pi_\mathrm{backup}(s), & \mathrm{otherwise}.
    \end{cases}
\end{align}

\section{Implementation}
\label{sec:implementation}

The presented algorithms, for estimating the aleatoric or epistemic uncertainty of an agent, are tested in simulated intersection scenarios in this study. However, these algorithms provide a general approach and could in principle be applied to any type of driving scenarios.
This section describes how the different test scenarios are set up, the MDP formulation of the decision-making problem, the design of the neural network architecture, and the details of the training process.

\subsection{Simulation setup}
\label{sec:simulation_setup}

Two occluded intersection scenarios are used in this study, shown in Fig.~\ref{fig:sparse_scenario} and~\ref{fig:dense_scenario}. The first scenario includes sparse traffic and aims to illustrate the qualitative difference between risk-neutral and risk-averse policies. The second scenario includes dense traffic and is used to compare the different algorithms, both qualitatively and quantitatively.
The scenarios were parameterized to create complicated traffic situations, where an optimal policy has to consider both the occlusions and the intentions of the other vehicles, sometimes drive through the intersection at a high speed, and sometimes wait at the intersection for an extended period of time. 

The Simulation of Urban Mobility (SUMO) was used to run the simulations~\cite{SUMO2018}. The controlled ego vehicle, a $12$ m long truck, aims to pass the intersection, in which it must yield to the crossing traffic.
In each episode, the ego vehicle starts $s_\mathrm{start}=200$ m south from the intersection, with its desired speed $v_\mathrm{set} = 15$ m/s. Passenger cars are randomly inserted into the simulation from the east and west end of the road network, with an average rate of $\rho_\mathrm{s}=0.1$ and $\rho_\mathrm{d}=0.5$ inserted vehicles per second for the sparse and dense traffic scenarios, respectively. The cars intend to either cross the intersection or turn to the right. The desired speed of the cars is uniformly distributed in the range $[v_\mathrm{min},v_\mathrm{max}] = [10, 15]$ m/s, and the longitudinal speed is controlled by the standard SUMO speed controller (which is a type of adaptive cruise controller, based on the intelligent driver model (IDM)~\cite{IDM}), with the exception that the cars ignore the presence of the ego vehicle. Normally, the crossing cars would brake to avoid a collision with the ego vehicle, even when the ego vehicle violates the traffic rules and does not yield. With this exception, however, more collisions occur, which gives a more distinct quantitative difference between different policies. Each episode is terminated when the ego vehicle has passed the intersection, when a collision occurs, or after $N_\mathrm{max}=100$ simulation steps. The simulations use a step size of $\Delta t = 1$ s.

Note that the setup of these scenarios includes two important sources of randomness in the outcome for a given policy, which the aleatoric uncertainty estimation should capture.
From the viewpoint of the ego vehicle, a crossing vehicle can appear at any time until the ego vehicle is sufficiently close to the intersection, due to the occlusions. Furthermore, there is uncertainty in the underlying driver state of the other vehicles, most importantly in the intention of going straight or turning to the right, but also in the desired speed.

Epistemic uncertainty is introduced by a separate test, in which the trained agent faces situations outside of the training distribution. In these test episodes, the maximum speed $v_\mathrm{max}$ of the surrounding vehicles are gradually increased from $15$ m/s (which is included in the training episodes) to $25$ m/s. To exclude effects of aleatoric uncertainty in this test, the ego vehicle starts in the non-occluded region close to the intersection, with a speed of $7$ m/s.

\subsection{MDP formulation}
The following Markov decision process describes the decision-making problem.

\subsubsection{State space, $\mathcal{S}$}
The state of the system,
\begin{align}
    s = (\{x_i, y_i, v_i, \psi_i\}_{i\in 0,\dots,N_\mathrm{veh}}),
\end{align}
consists of the position $x_i$, $y_i$, longitudinal speed $v_i$, and heading $\psi_i$, of each vehicle, where index $0$ refers to the ego vehicle. The agent that controls the ego vehicle can observe other vehicles within the sensor range $x_\mathrm{sensor}=200$ m, unless they are occluded.

\subsubsection{Action space, $\mathcal{A}$}

At every time step, the agent can choose between three high-level actions: \textit{`stop'}, \textit{`cruise'}, and \textit{`go'}, which are translated into accelerations through the IDM. 
The action \textit{`go'} makes the IDM control the speed towards $v_\mathrm{set}$ by treating the situation as if there are no preceding vehicles, whereas \textit{`cruise'} simply keeps the current speed. The action \textit{`stop'} places an imaginary target vehicle just before the intersection, which causes the IDM to slow down and stop at the stop line. If the ego vehicle has already passed the stop line, \textit{`stop'} is interpreted as maximum braking. Finally, the output of the IDM is limited to $[a_\mathrm{min},a_\mathrm{max}] = [\minus3, 1]$ m/s\textsuperscript{2}. Note that the agent takes a new decision at every time step $\Delta t$ and can therefore switch between, e.g., \textit{`stop'} and \textit{`go'} multiple times during an episode.

\subsubsection{Reward model, $R$}
The objective of the agent is to drive through the intersection in a time efficient way, without colliding with other vehicles. A simple reward model is used to achieve this objective. The agent receives a positive reward $r_\mathrm{goal}=10$ when the ego vehicle manages to cross the intersection and a negative reward $r_\mathrm{col}=\minus10$ if a collision occurs. If the ego vehicle gets closer to another vehicle than $2.5$ m longitudinally or $1$ m laterally, a negative reward $r_\mathrm{near}=\minus10$ is given, but the episode is not terminated. At all other time steps, the agent receives $0$ as reward.

\subsubsection{Transition model, $T$}
The state transition probabilities are not known by the agent, and they are implicitly defined by the simulation model, described in Sect.~\ref{sec:simulation_setup}.

\subsection{Backup policy}
A simple backup policy $\pi_\mathrm{backup}(s)$ is used together with the uncertainty criteria. This policy selects the action \textit{`stop'} if the vehicle is able to stop before the intersection, considering the braking limit $a_\mathrm{min}$. Otherwise, the backup policy selects the action that is recommended by the agent. If the backup policy would always consist of \textit{`stop'}, the ego vehicle could end up standing still in the intersection and thereby cause more collisions.
Naturally, more advanced backup policies would be considered in a real-world implementation, for example based on optimal control~\cite{Svensson2018}, but such a policy would not significantly change the results of this study.

\subsection{Neural network architecture}

In previous work, we introduced a one-dimensional convolutional neural network architecture, which improves both the training speed and final performance, compared to a standard fully connected architecture~\cite{Hoel2018}. By applying convolutional layers and a maxpooling layer to the input that describes the state of the surrounding vehicles, the output becomes both invariant to the ordering of the surrounding vehicles in the input vector and independent of the number of surrounding vehicles. A more detailed description of this architecture is provided in the previous work~\cite{Hoel2018}.

\begin{figure}[!t]
    \centering
    \includegraphics[width=0.98\columnwidth]{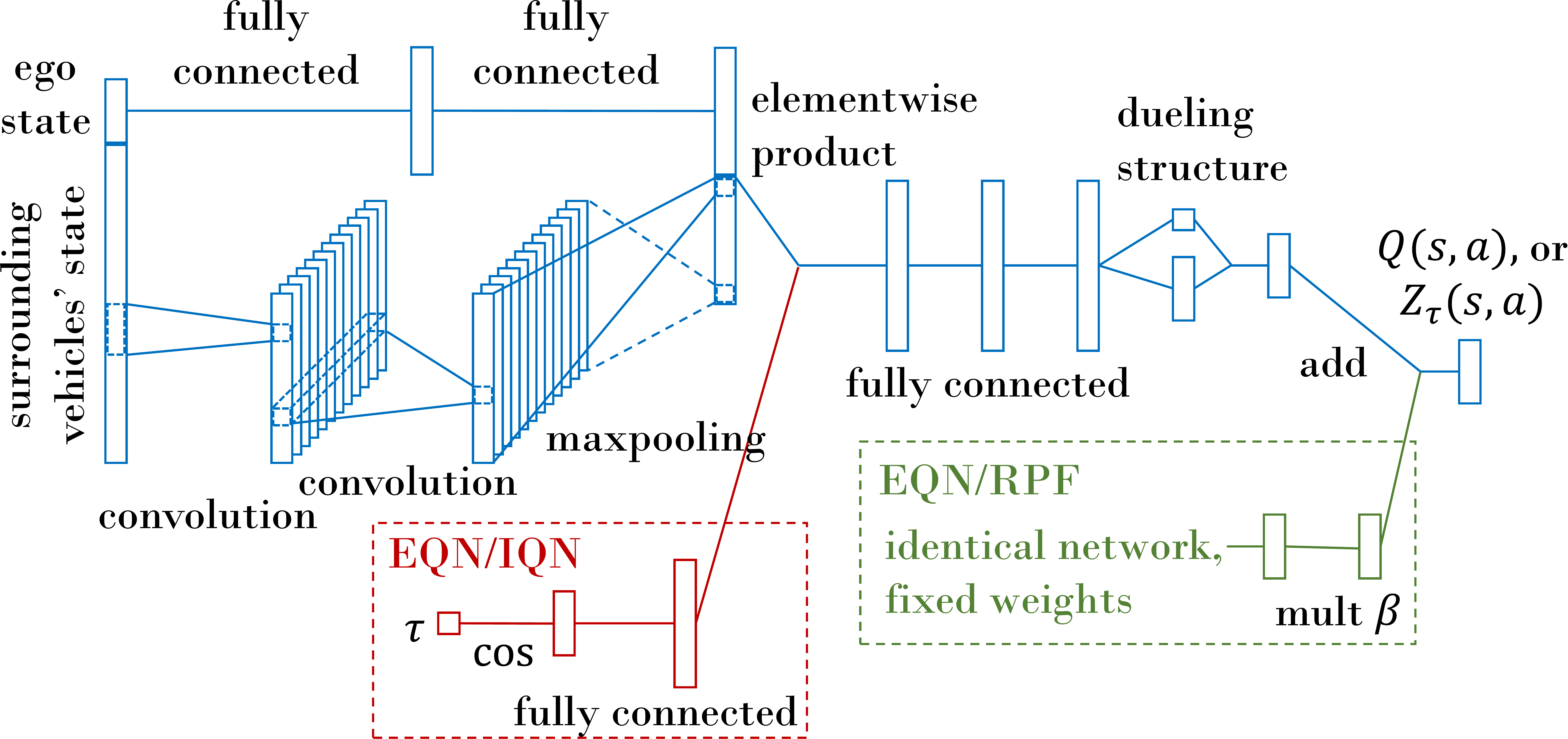}
    \caption{The neural network architecture that is used for the different agents. The red part is included for the EQN and IQN agents, whereas the green part is included for the EQN and RPF agents.}
    \label{fig:network_architecture}
\end{figure}

Fig.~\ref{fig:network_architecture} shows the neural network architecture that is used in this study. The size and stride of the first convolutional layers are set to four, which is equal to the number of states that describe each surrounding vehicle, whereas the second convolutional layer has a size and stride of one. Both convolutional layers have $256$ filters each, and all fully connected layers have $256$ units. Finally, a dueling structure~\cite{Wang2016}, which separates the estimation of the value of a state and the advantage of an action, outputs $Z_\tau(s,a)$ or $Q(s,a)$, depending on which algorithm that is used. All layers use rectified linear units (ReLUs) as activation functions, except for the dueling layer, which has a linear activation function. Before the state $s$ is fed to the network, each entry is normalized to the range $[-1,1]$ by considering the possible minimum and maximum values.

The network architecture for the IQN agent has an additional input for the sample quantile $\tau$, shown in Fig.~\ref{fig:network_architecture}. As proposed by Dabney et al.~\cite{Dabney2018}, an embedding from $\tau$ is created by setting $\phi(\tau) = (\phi_1(\tau), \ldots, \phi_{64}(\tau))$, where $\phi_j(\tau) = \cos{(\pi j \tau)}$, and then passing $\phi(\tau)$ through a fully connected layer with $512$ units. The output of the embedding is then merged with the output of the concatenating layer as the element-wise (Hadamard) product.

\begin{table}[t]
	\renewcommand{\arraystretch}{1.2}
	\caption{Hyperparameters of the different algorithms}
	\label{tab:hyperparameters}
	\centering
	\begin{tabular}{l|lr}
		\toprule
		IQN, & Number of quantile samples, $N, N', K_\tau$ & $32$\\
		EQN & CVaR parameter, $\alpha$ & $1$\\
		
		\midrule
		
		RPF, & Number of ensemble members, $K$ & $10$\\
		EQN & Prior scale factor, $\beta$ & $300$\\
		& Experience adding probability, $p_\mathrm{add}$ & $0.5$\\
		
		\midrule
		
		DQN, & Discount factor, $\gamma$ & $0.95$\\
		IQN, & Learning start iteration, $N_\mathrm{start}$ & $50{,}000$\\
		RPF, & Replay memory size, $N_\mathrm{replay}$ & $500{,}000$\\
		EQN & Learning rate, $\eta$ & $0.0005$\\
		& Mini-batch size, $|M|$ & $32$\\
		& Target network update frequency, $N_\mathrm{update}$ & $20{,}000$\\
		& Huber loss threshold, $\kappa$ & $10$\\

		\midrule

		DQN, & Initial exploration parameter, $\epsilon_0$  & $1$\\
		IQN & Final exploration parameter, $\epsilon_1$ & $0.05$\\
		& Final exploration iteration, $N_\epsilon$ & $500{,}000$\\

		\bottomrule
	\end{tabular}
\end{table}

\subsection{Training process}
Algorithm \ref{alg:iqn_training}, \ref{alg:rpf_training}, and~\ref{alg:eqn_training} were used to train the IQN, RPF, and EQN agents, respectively. Additionally, the Double DQN trick was used to reduce overestimation of the \mbox{$Q$-values}~\cite{Hasselt2016}, which subtly changes the maximization part in Eq.~\ref{eq:td_dqn} and~\ref{eq:loss_rpf}.
During the training of the DQN and IQN agents, an \mbox{$\epsilon$-greedy} exploration policy was followed, where $\epsilon$ was linearly decreased from $\epsilon_0$ to $\epsilon_1$ over $N_\epsilon$ training steps. Huber loss was applied to the TD-error of all the algorithms, in order to improve the robustness of the training process, and the neural network weights were updated by the Adam optimizer~\cite{Kingma2014}. The training process was parallelized for the ensemble-based versions, in order to reduce the training time. 
Table~\ref{tab:hyperparameters} displays the hyperparameters that were used for the different algorithms.
Due to the computational complexity, a systematic grid search was not performed. Instead, the hyperparameter values were selected from an informal search, based upon the values given by Mnih et al.~\cite{Mnih2015}, Dabney et al.~\cite{Dabney2018}, and Osband et al.~\cite{Osband2018}. Additional results are also presented for a set of different values of $\alpha$, $\beta$, and $K$, in order to demonstrate how the choice of these parameters influence the behavior of the agent.

As mentioned in Sect.~\ref{sec:simulation_setup}, an episode is terminated due to a timeout after maximally $N_\mathrm{max}$ steps, since otherwise the current policy could make the ego vehicle stop at the intersection indefinitely. However, since the time is not part of the state space, a timeout terminating state is not described by the MDP. 
Therefore, in order to make the agents act as if the episodes have no time limit, the last experience of a timeout episode is not added to the experience replay buffer.

All the agents are trained for $3{,}000{,}000$ training steps, at which point the agents' policies have converged, and then the trained agents are tested on $1{,}000$ randomly initialized test episodes. The test episodes are generated in the same way as the training episodes, described in Sect.~\ref{sec:simulation_setup}, but they are not present during the training phase. Furthermore, the set of test episodes is identical for all the trained agents, in order to provide an appropriate comparison. Each agent is trained with five random seeds and the mean results are presented, together with the corresponding standard deviation.

\section{Results}
\label{sec:results}

The results show that the IQN method can be used to estimate the aleatoric uncertainty in a traffic situation and the uncertainty criterion can be used to identify situations with high uncertainty, in order to prevent collisions. The results also illustrate that the ensemble RPF method can provide an estimate of the epistemic uncertainty and use the uncertainty criterion to classify situations as within or outside the training distribution. Furthermore, the results of the EQN method demonstrate that this approach provides a complete estimate of both types of uncertainty. This section presents the results in detail, together with an analysis of the characteristics of the results, whereas a broader discussion on the properties of the algorithms follows in Sect.~\ref{sec:discussion}. 
Animations of the presented scenarios are available on GitHub~\cite{sourceCode2021}.

\subsection{Aleatoric uncertainty estimation}

\begin{figure}[!t]
    \centering
    \includegraphics[width=0.98\columnwidth]{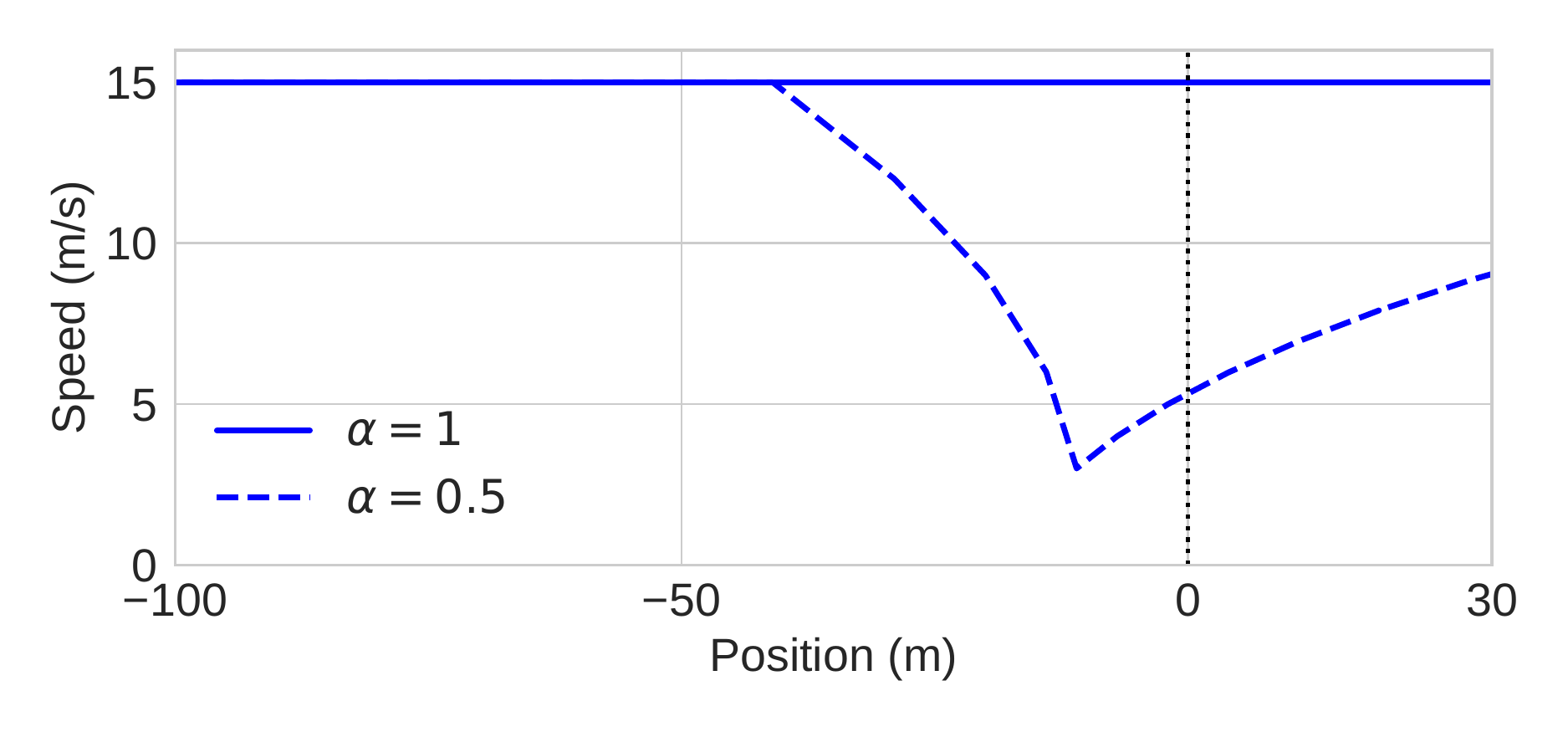}
    \caption{Speed of the ego vehicle as a function of distance to the occluded intersection, positioned at the dotted vertical line, for a sparse traffic scenario. In this episode, no crossing vehicles are observed.}
    \label{fig:sparse_speed}
\end{figure}

\begin{figure}[!t]
    \centering
    \subfloat[The ego vehicle is shown in red, the occluded vehicle in yellow, and the areas that cause occlusions are displayed in gray.]
    {\includegraphics[width=0.98\columnwidth]{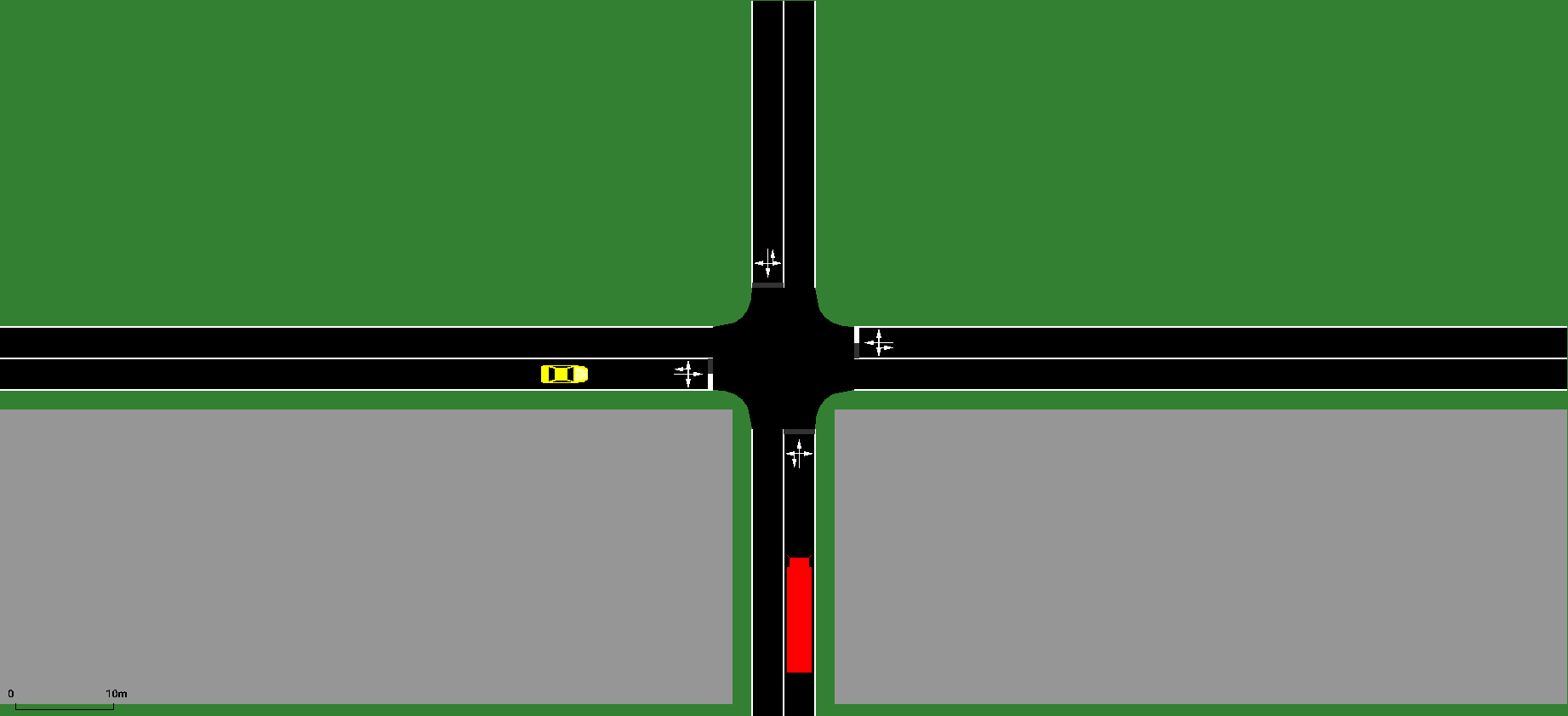}
    \label{fig:sparse_scenario}}
    \hfil
    \subfloat[$\alpha = 1$]
    {\includegraphics[width=0.48\columnwidth]{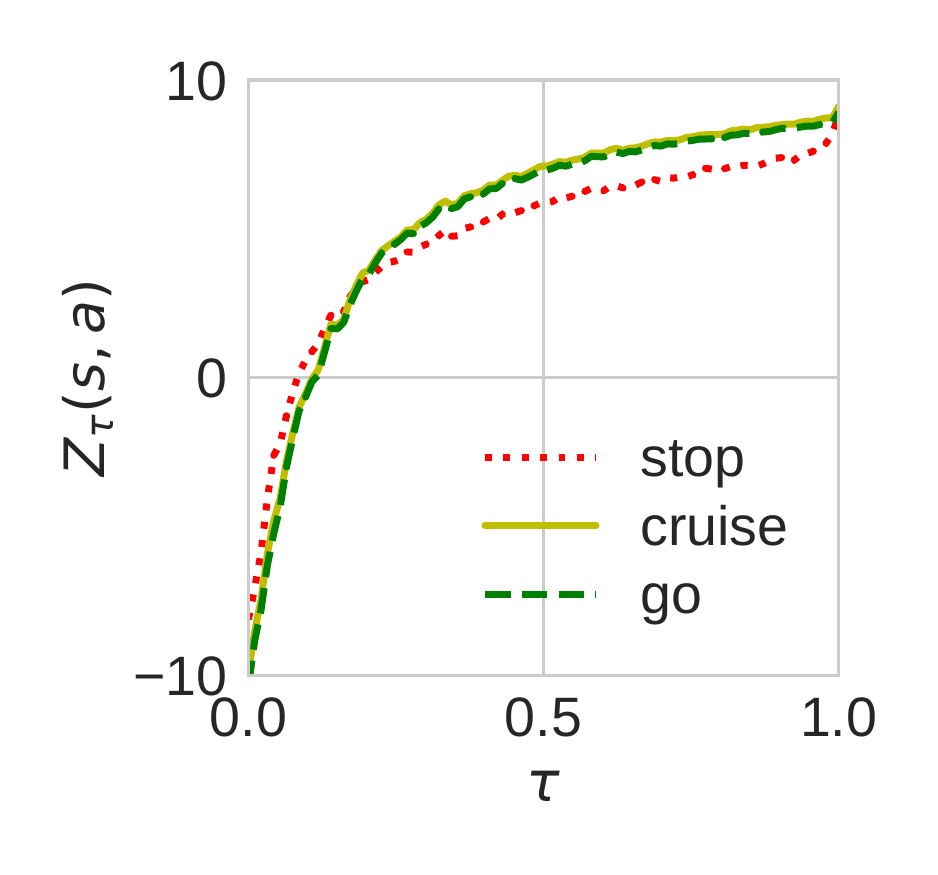}
    \label{fig:sparse_quantiles_cvar_1}}
    \hfil
    \subfloat[$\alpha = 0.5$]
    {\includegraphics[width=0.48\columnwidth]{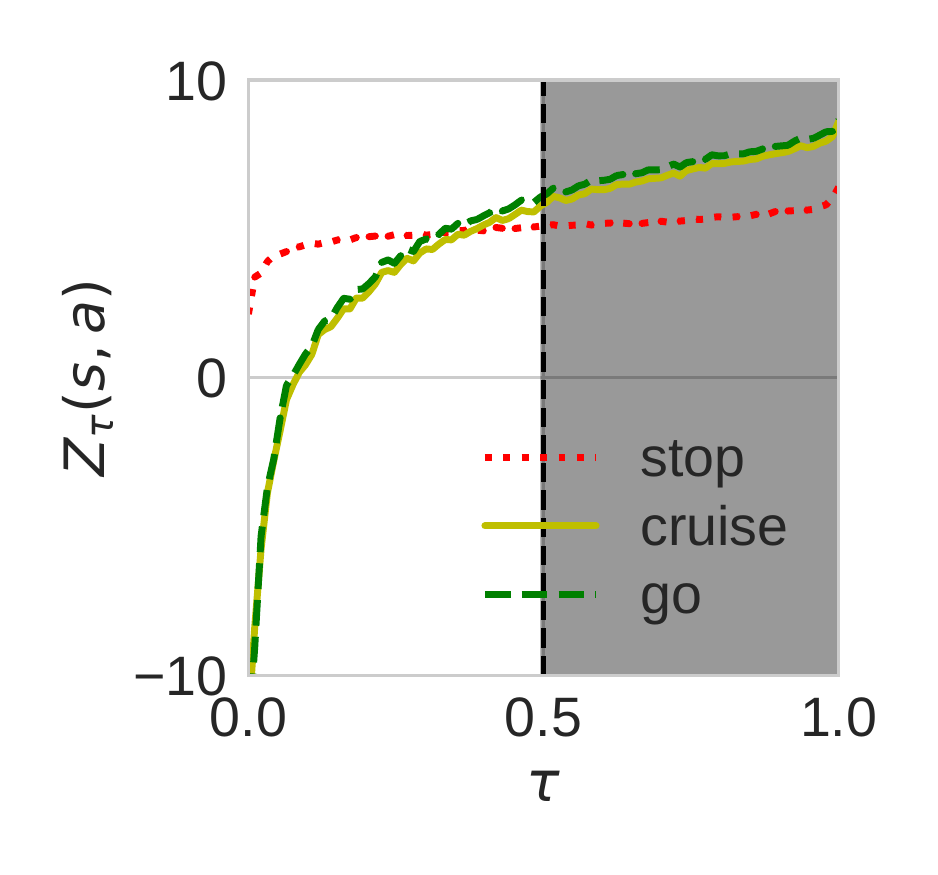}
    \label{fig:sparse_quantiles_cvar_0p5}}
    \caption{Example of a situation that results in a collision for an agent with risk-neutral training ($\alpha=1$) but is solved without collisions with risk-sensitive training ($\alpha=0.5$). The estimated quantile function of the random variable $Z_\mathrm{\tau}(s,a)$ is shown for both cases and indicates that both agents are aware of the aleatoric uncertainty in the situation.}
    \label{fig:sparse_state}
\end{figure}

To illustrate the behavior of the trained IQN agent and provide intuition on how risk-sensitive training affects the obtained policy, results for the sparse traffic scenario are first displayed. Table~\ref{tab:sparse_scenario} shows nearly identical quantitative results for a trained risk-neutral IQN agent and a DQN agent. Both agents find a policy that drives through the intersection at the maximum speed if no crossing vehicles are observed, see Fig.~\ref{fig:sparse_speed}. Since crossing traffic is sparse, this policy maximizes the expected return, but causes collisions in around one out of ten test episodes. An IQN agent that is trained in a risk-sensitive way, by setting the CVaR parameter $\alpha=0.5$, instead slows down and passes the occluded area with a low speed, which allows the ego vehicle to stop before the intersection if a crossing vehicle appears. Such a policy solves almost all test episodes without collisions, but increases the mean duration of an episode, hereafter referred to as crossing time, with around $50\%$.
An example of a situation that causes a collision with risk-neutral training but is collision-free with risk-sensitive training, is shown in Fig.~\ref{fig:sparse_scenario}. In this situation, the ego vehicle is driving at $15$ m/s, while an occluded vehicle is approaching from the west. Fig.~\ref{fig:sparse_quantiles_cvar_1} and~\ref{fig:sparse_quantiles_cvar_0p5} display the estimated quantile function of the return distribution $Z_\tau(s,a)$, which reveal that both agents are aware of the risk of a collision, indicated by the small probability ($\tau < 0.1$) of a negative return.
However, different policies are obtained, due to the difference in risk-sensitivity. The reason for the high aleatoric uncertainty in actions \textit{`go'} and \textit{`cruise'} of the risk-averse agent (Fig.~\ref{fig:sparse_quantiles_cvar_0p5}) is that the agent will not be able to later decide to stop before the intersection, due to the limited braking capacity.

\begin{table}[t]
	\renewcommand{\arraystretch}{1.2}
	\caption{Sparse traffic scenario}
	\label{tab:sparse_scenario}
	\centering
	\begin{tabular}{llcc}
		\toprule
		&& collisions (\%) & crossing time (s)\\
		\midrule
		IQN & $\alpha = 1$  & $10.8 \pm 0.2$ & $15.8 \pm 0.1$\\
		IQN & $\alpha = 0.5$ & $0.1 \pm 0.1$ & $24.0 \pm 0.4$\\
		DQN & & $10.7 \pm 0.2$ & $15.9 \pm 0.1$\\
		\bottomrule
	\end{tabular}
\end{table}

\begin{figure}[!t]
    \centering
    \includegraphics[width=0.98\columnwidth]{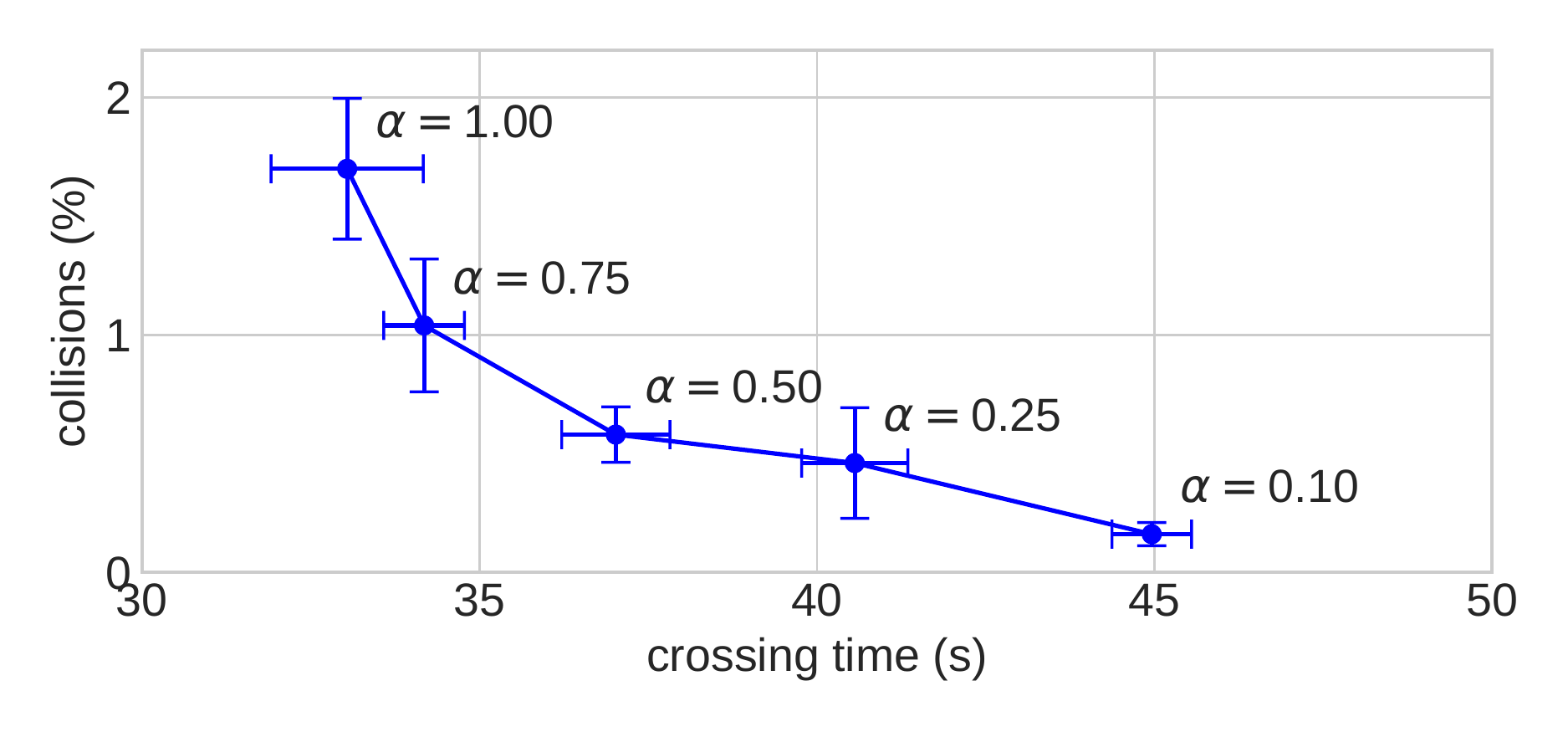}
    \caption{Number of collisions and crossing time for different levels of risk-sensitive training, which is achieved by varying the CVaR parameter $\alpha$.}
    \label{fig:pareto_alpha}
\end{figure}

Table~\ref{tab:main_scenario} shows how the trained IQN agent performs in the second test scenario, in which traffic is dense and the occluding objects are placed further from the intersection. The results illustrate the natural trade-off between time efficiency and safety. With a more risk-averse training (lower value of the CVaR parameter $\alpha$), the number of collisions is reduced, but the time it takes to cross the intersection increases, see Fig.~\ref{fig:pareto_alpha}.

\begin{figure}[!t]
    \centering
    \includegraphics[width=0.98\columnwidth]{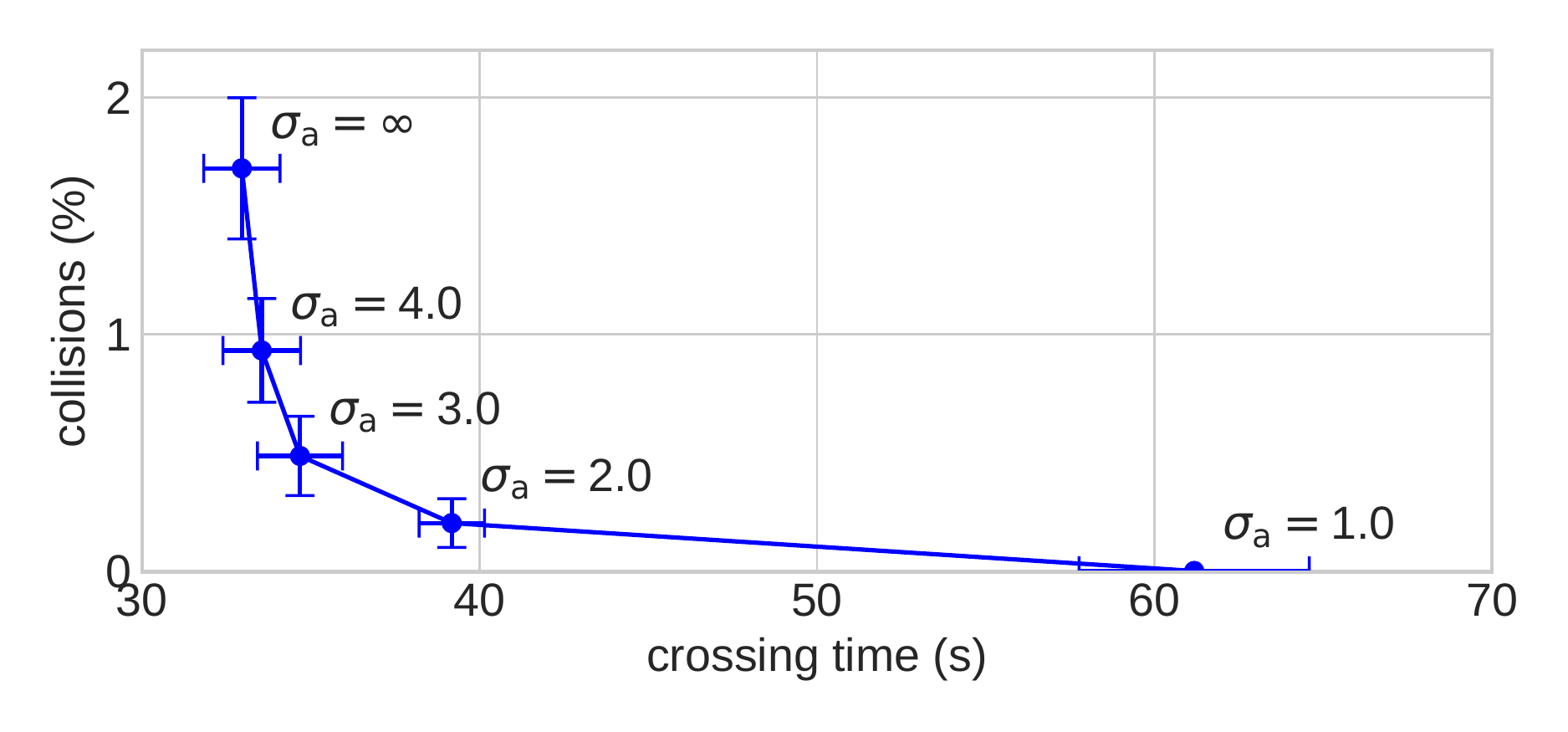}
    \caption{Number of collisions and crossing time for the IQN algorithm for different levels of allowed aleatoric uncertainty, which is achieved by varying the parameter $\sigma_\mathrm{a}$.}
    \label{fig:pareto_sigma_a}
\end{figure}

\begin{figure}[!t]
    \centering
    \subfloat[The ego vehicle is shown in red and has a speed of $1$ m/s.]
    {\includegraphics[width=0.98\columnwidth]{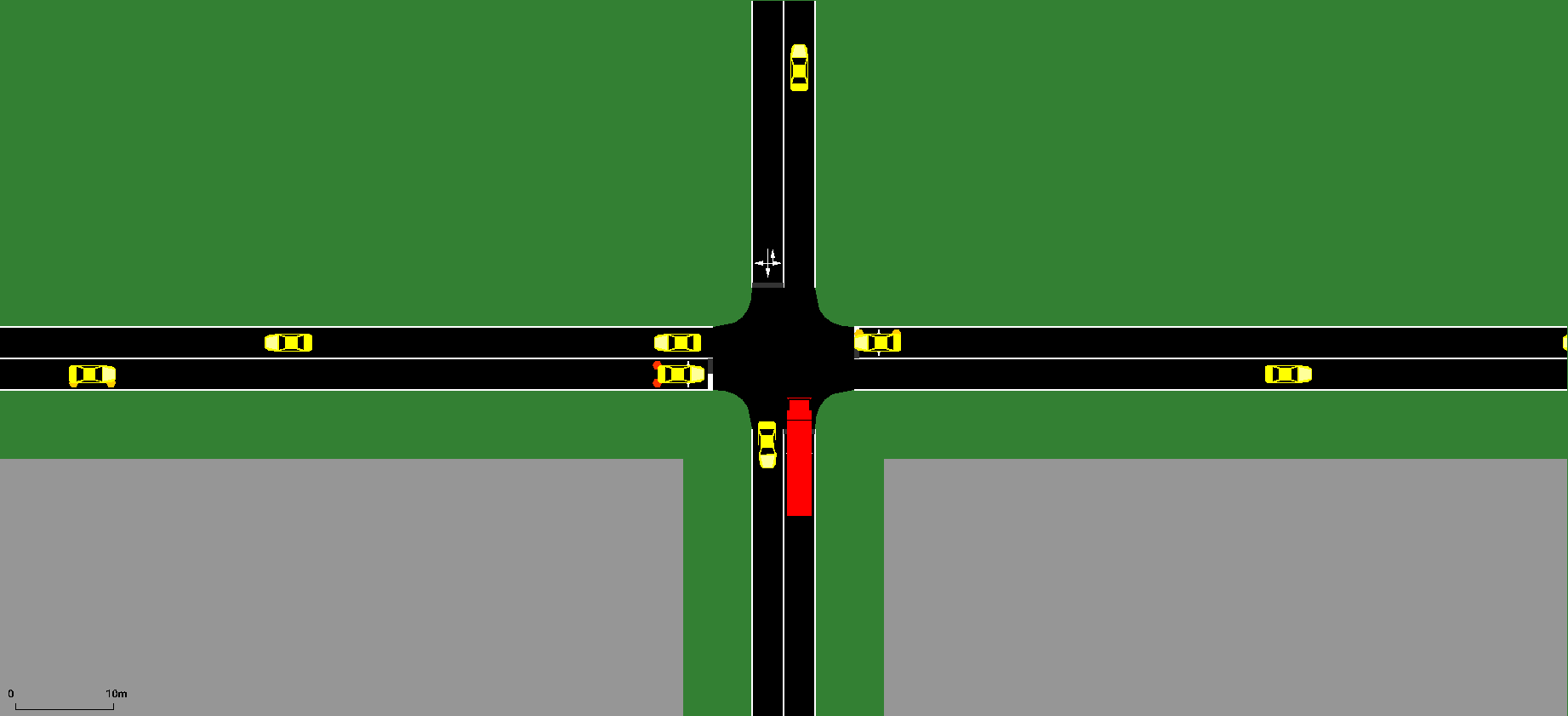}
    \label{fig:dense_scenario}}
    \hfil
    \subfloat[Estimated quantile function of the random variable $Z_\mathrm{\tau}(s,a)$.]
    {\includegraphics[width=0.98\columnwidth]{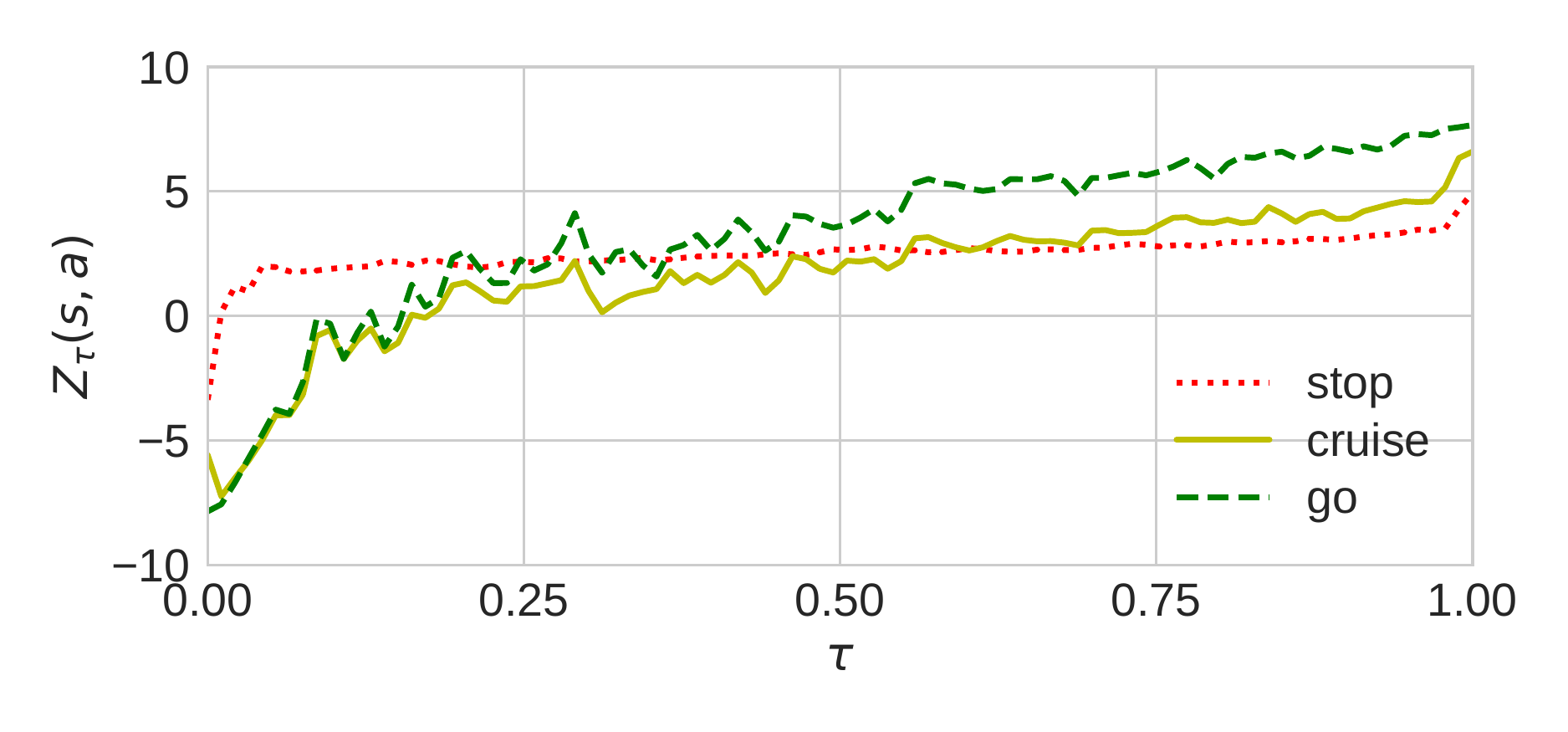}
    \label{fig:situation_sigma_a_quantiles}}
    \caption{Example of a situation with high aleatoric uncertainty for the actions \textit{`go'} and \textit{`cruise'}, due to uncertainty in the intention of the vehicle that is about to enter the intersection from the west.}
    \label{fig:situation_sigma_a}
\end{figure}

Furthermore, the results in Table~\ref{tab:main_scenario} and Fig.~\ref{fig:pareto_sigma_a} demonstrate that the IQN method, combined with the aleatoric uncertainty criterion, can be used to detect situations with high aleatoric uncertainty. When the maximum allowed uncertainty is reduced (lower values of $\sigma_\mathrm{a}$), the number of collisions is reduced. However, similarly to training in a risk-averse manner, a more conservative policy increases the time it takes to cross the intersection. An example situation with high aleatoric uncertainty, due to uncertainty in the intention of another vehicle, is shown in Fig.~\ref{fig:dense_scenario}. The car that is approaching the intersection from the west has here slowed down due to a preceding car, which turned to the south. Due to the low speed, the IQN agent expects that the approaching car will also turn to the south. Therefore, the agent estimates that in most cases it would be best to choose the action \textit{`go'}, to immediately cross the intersection, see Fig.~\ref{fig:situation_sigma_a_quantiles}. However, the agent also estimates that with a low probability, this action can cause a collision, which is indicated by the negative values of the estimated return distribution $Z_\mathrm{\tau}$. Since the sample variance is high in this situation, $\mathrm{Var}_{\tau_\sigma}[Z_\mathrm{\tau}(s,a_\mathrm{go})]=12.0$, an uncertainty criterion with $\sigma_\mathrm{a}^2 < 12.0$ prevents a collision by choosing the backup policy, i.e., stopping at the intersection.

\begin{table}[t]
	\renewcommand{\arraystretch}{1.2}
	\caption{Dense traffic scenario, tested within the training distribution}
	\label{tab:main_scenario}
	\centering
	\begin{tabular}{l|lcc}
		\toprule
		algorithm, & variable & collisions (\%) & crossing time (s)\\
		fixed param. & parameter & & \\
		\midrule
		DQN & - & $4.0 \pm 0.5$ & $31.7 \pm 1.1$\\
		\midrule
		IQN, & $\alpha = 1.0$ & $1.7 \pm 0.3$ & $33.0 \pm 1.1$\\ 
		$\sigma_\mathrm{a}=\infty$ & $\alpha = 0.75$ & $1.0 \pm 0.3$ & $34.2 \pm 0.6$\\
		& $\alpha = 0.5$ & $0.6 \pm 0.1$ & $37.0 \pm 0.8$\\
		& $\alpha = 0.25$ & $0.4 \pm 0.2$ & $40.6 \pm 0.8$\\
		& $\alpha = 0.1$ & $0.2 \pm 0.1$ & $45.0 \pm 0.6$\\
		\midrule
		IQN, & $\sigma_\mathrm{a} = \infty$ & $1.7 \pm 0.3$ & $33.0 \pm 1.1$\\ 
		$\alpha=1$ & $\sigma_\mathrm{a} = 4.0$ & $0.9 \pm 0.2$ & $33.5 \pm 1.2$\\
		& $\sigma_\mathrm{a} = 3.0$ & $0.5 \pm 0.2$ & $34.7 \pm 1.3$\\
		& $\sigma_\mathrm{a} = 2.0$ & $0.2 \pm 0.1$ & $39.2 \pm 1.0$\\
		& $\sigma_\mathrm{a} = 1.0$ & $0.0 \pm 0.0$ & $61.2 \pm 3.4$\\
		\midrule
		RPF, & $\beta=0$ & $3.0 \pm 0.2$ & $29.4 \pm 0.3$\\
		$K=10$ & $\beta=100$ & $2.8 \pm 0.4$ & $32.1 \pm 0.5$\\
		& $\beta=300$ & $1.5 \pm 0.3$ & $38.0 \pm 1.8$\\   
		& $\beta=1000$ & $1.8 \pm 0.4$ & $44.6 \pm 1.0$\\
		\midrule
		RPF, & $K=3$ & $3.0 \pm 1.0$ & $34.8 \pm 1.6$\\
		$\beta=300$ & $K=10$ & $1.5 \pm 0.3$ & $38.0 \pm 1.8$\\   
		& $K=30$ & $1.9 \pm 0.4$ & $34.6 \pm 1.4$\\
		\midrule
		EQN, & $\sigma_\mathrm{a} = \infty$ & $0.9 \pm 0.1$ & $32.0 \pm 0.2$\\
		$\alpha = 1.0$, & $\sigma_\mathrm{a} = 3.0$ & $0.6 \pm 0.2$ & $33.8 \pm 0.3$\\
		$K=10$, & $\sigma_\mathrm{a} = 2.0$ & $0.5 \pm 0.1$ & $38.4 \pm 0.5$\\
		$\beta = 300$ & $\sigma_\mathrm{a} = 1.5$ & $0.3 \pm 0.1$ & $47.2 \pm 1.2$\\
		& $\sigma_\mathrm{a} = 1.0$ & $0.0 \pm 0.0$ & $71.1 \pm 1.9$\\
		\cmidrule{2-4}
		& \begin{tabular}{@{}c@{}}$\sigma_\mathrm{a} = 1.5$, \\ $\sigma_e=1.0$ \end{tabular} & $0.0 \pm 0.0$ & $48.9\pm 1.6$\\
		\bottomrule
	\end{tabular}
\end{table}

\subsection{Epistemic uncertainty estimation}

\begin{figure}[!t]
    \centering
    \includegraphics[width=0.98\columnwidth]{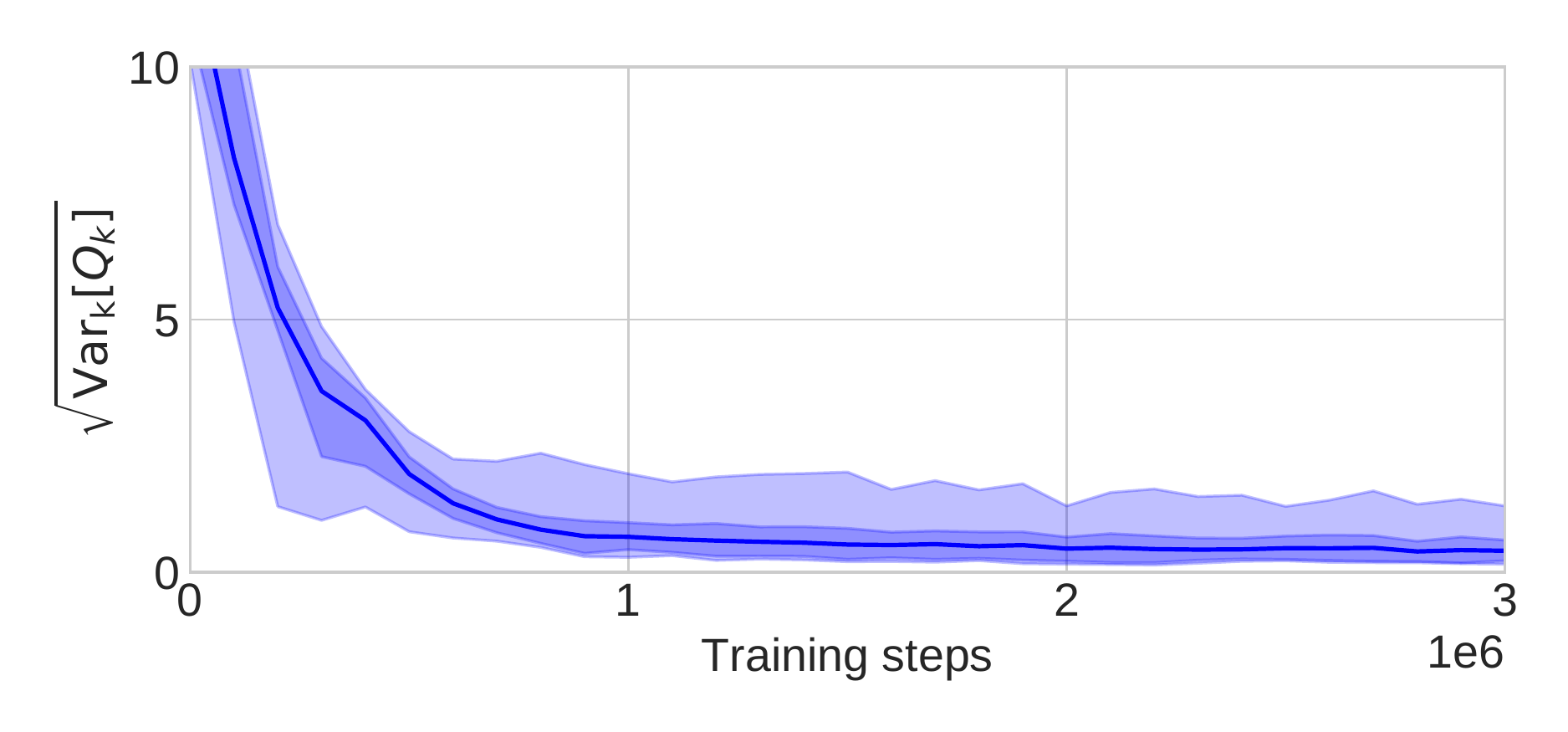}
    \caption{Epistemic uncertainty of the chosen actions for the ensemble RPF agent, with parameters $\beta=300$ and $K=10$, during testing episodes within the training distribution. The solid line shows the mean, while the shaded regions indicate percentile $10$ to $90$ and $1$ to $99$.}
    \label{fig:epistemic_uncertainty}
\end{figure}

The trained RPF agent performs similarly as the risk-neutral IQN agent in the dense traffic scenario, see Table~\ref{tab:main_scenario}. The parameter $\beta$, which scales the importance of the random prior network and thereby influences the exploration strategy, has a relatively small effect on the performance within the training distribution. Even setting $\beta=0$, which completely removes the effect of the random prior network and only relies on statistical bootstrapping for exploration, gives reasonable results. Similarly, the number of networks $K$ have a low effect on the performance. Fig.~\ref{fig:epistemic_uncertainty} shows how the epistemic uncertainty of the chosen actions during the test episodes is reduced during the training process.

\begin{figure}[!t]
    \centering
    \subfloat[$t=0$, {$\mathrm{Var}_k[Q_k(s,a_\mathrm{go})] = 57.8$}]
    {\includegraphics[width=0.98\columnwidth]{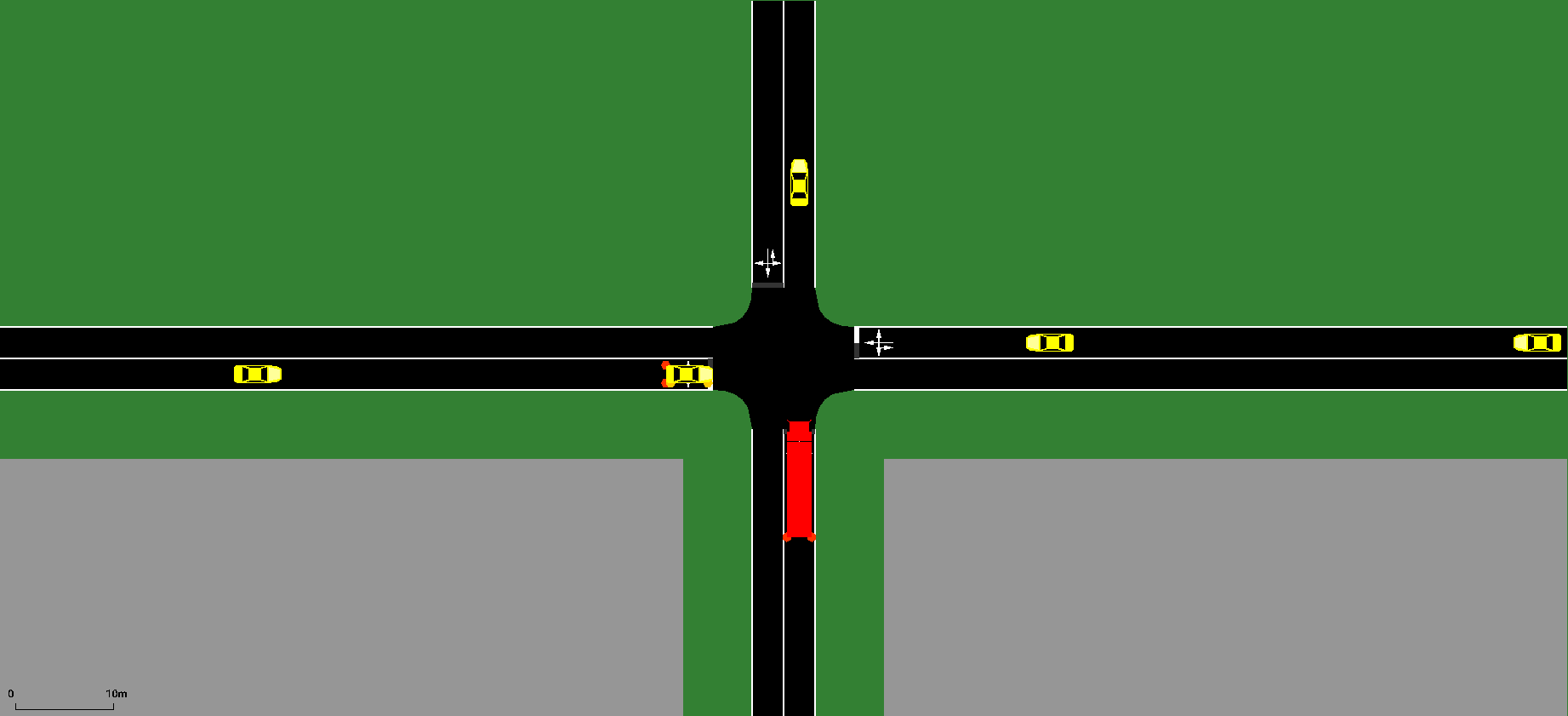}}
    \hfil
    \subfloat[$t=3$ s, $\sigma_\mathrm{e}=\infty$]
    {\includegraphics[width=0.48\columnwidth]{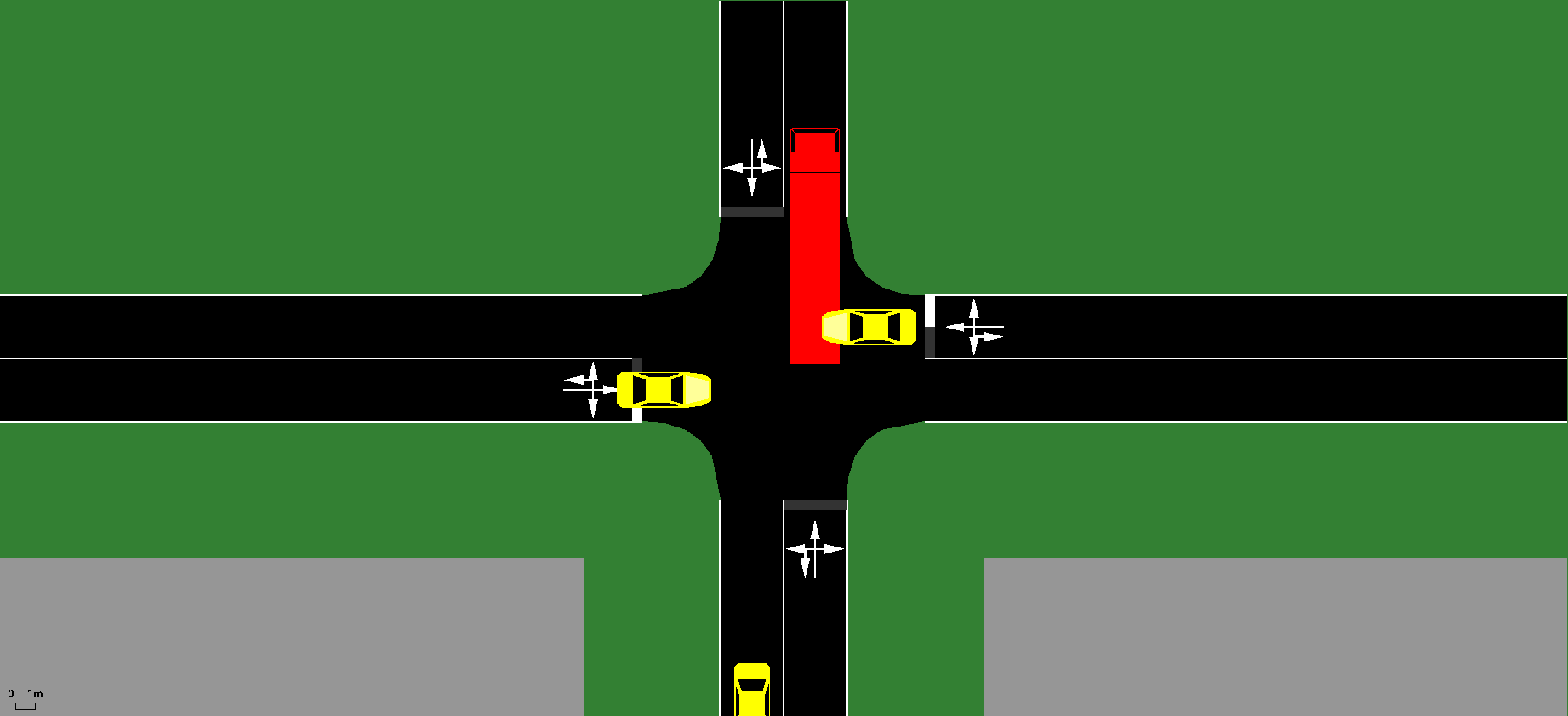}}
    \hfil
    \subfloat[$t=3$ s, $\sigma_\mathrm{e}=4$]
    {\includegraphics[width=0.48\columnwidth]{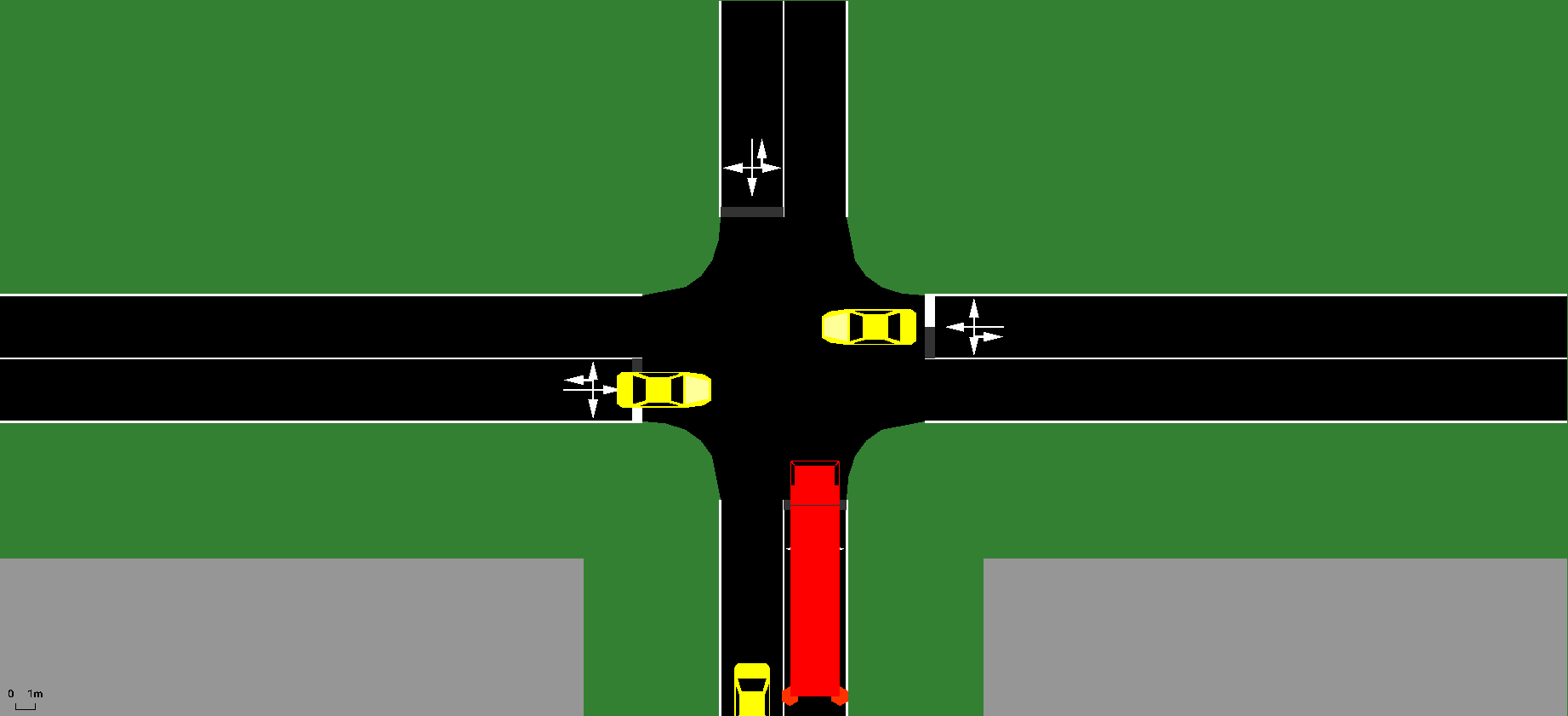}}
    \caption{Example of a situation with high epistemic uncertainty~(a), in which the eastmost vehicle approaches the intersection with a speed of $23$ m/s. Without the epistemic uncertainty threshold, the RPF agent chooses to cross the intersection, which causes a collision~(b), whereas the collision is avoided when the uncertainty criterion is applied~(c).}
    \label{fig:increased_speed_situation}
\end{figure}

\begin{figure}[!t]
    \centering
    \subfloat[Collisions]
    {\includegraphics[width=0.98\columnwidth]{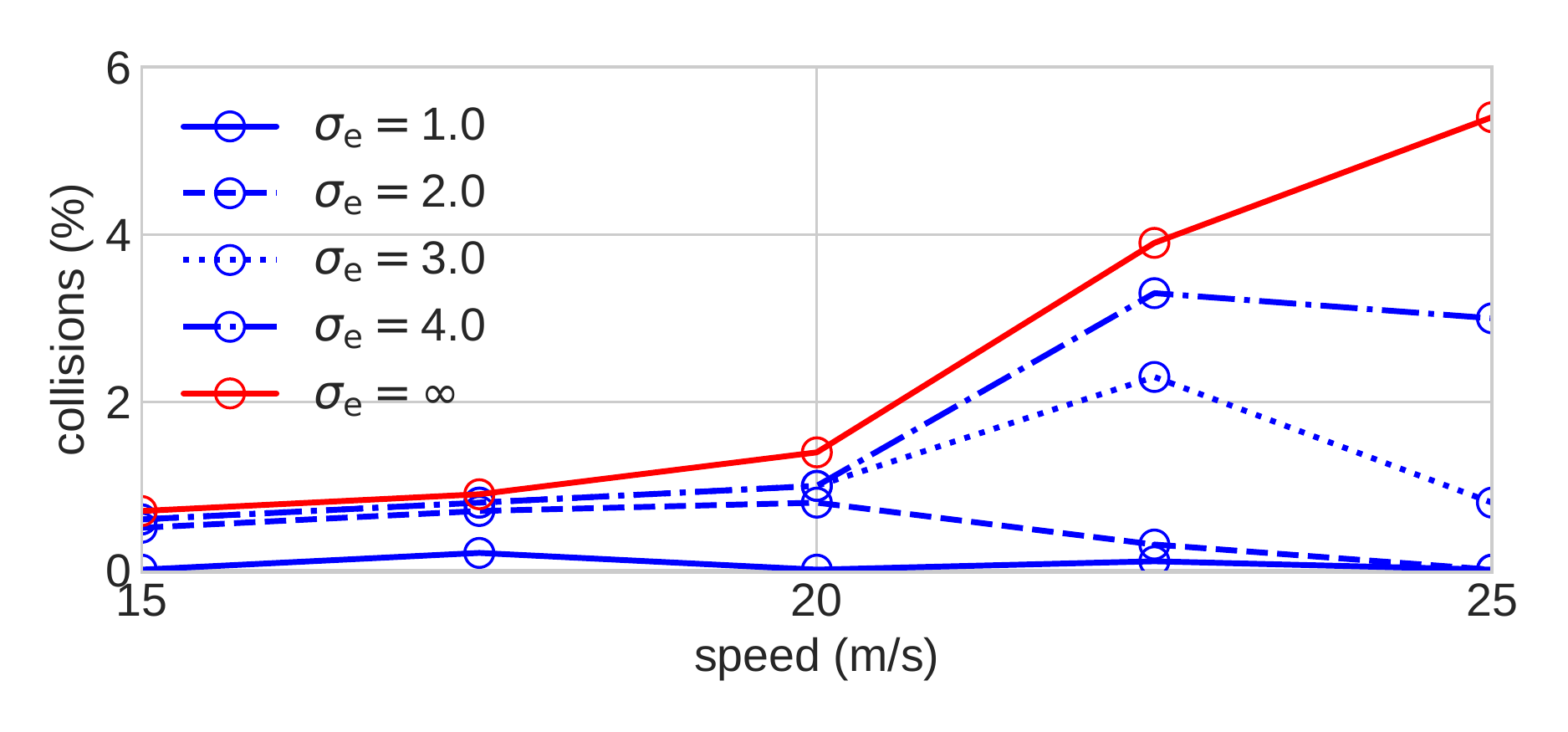}}
    \hfil
    \subfloat[Timeouts]
    {\includegraphics[width=0.98\columnwidth]{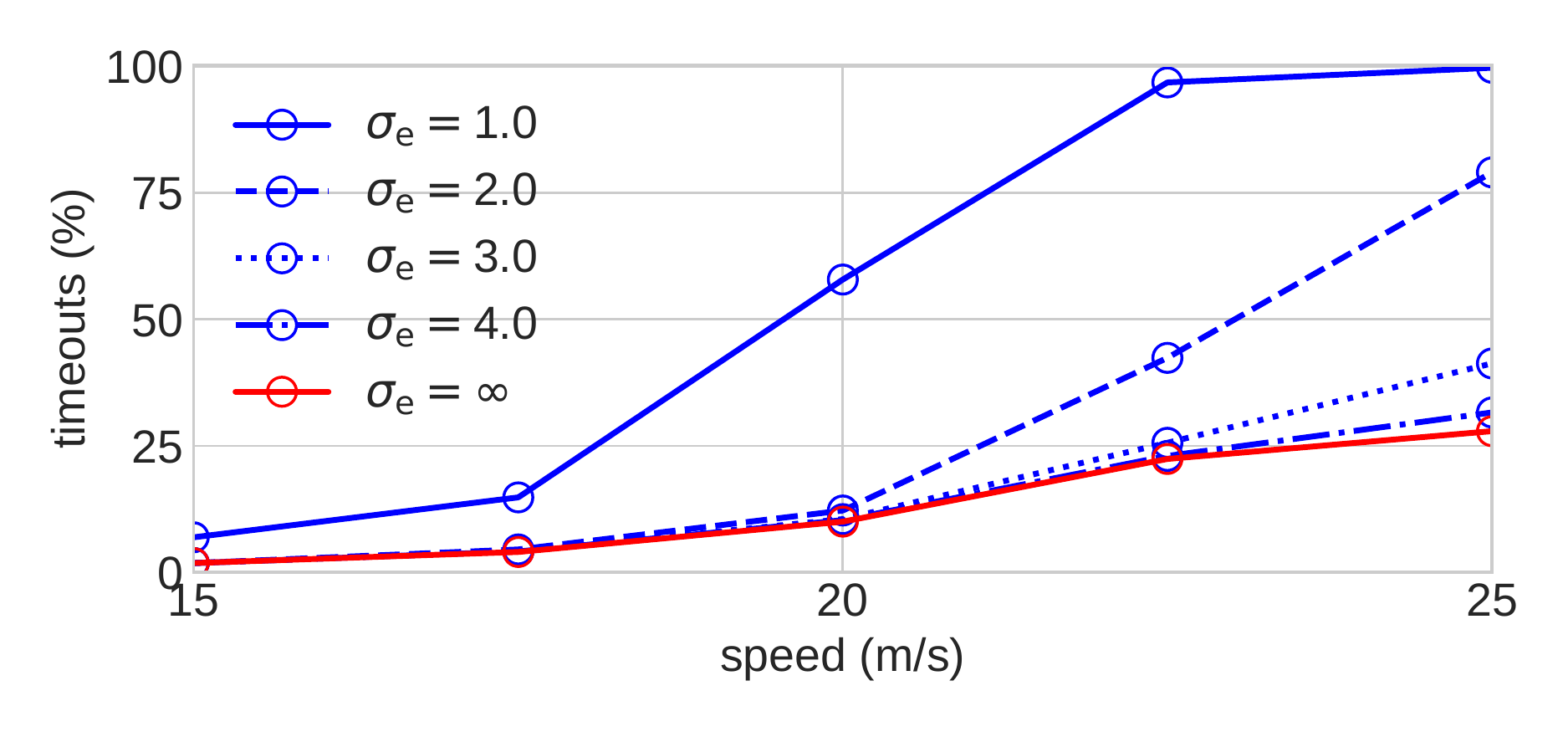}}
    \caption{Number of collisions and timeouts for the RPF agent ($\beta=300$, $K=10$), in situations outside the training distribution. The maximum speed of the crossing vehicles is $15$ m/s during the training process, and then the speed is gradually increased in the testing episodes.}
    \label{fig:increased_speed}
\end{figure}

To illustrate that the RPF agent can estimate the epistemic uncertainty and detect situations that are outside of the training distribution, the trained agent is exposed to crossing traffic with a higher speed than during the training episodes, see Sect.~\ref{sec:simulation_setup}.
Fig.~\ref{fig:increased_speed} shows that if no epistemic uncertainty threshold is used, i.e., setting $\sigma_\mathrm{e}=\infty$, the number of collisions increases significantly when the speed of the crossing vehicles increases. If the threshold $\sigma_\mathrm{e}$ is reduced, the number of collisions is reduced to almost zero, whereas the number of timeouts increases. An example of a situation with high epistemic uncertainty, in which a collision is avoided by limiting the allowed uncertainty, is shown in Fig.~\ref{fig:increased_speed_situation}.

The result at $15$ m/s, which is included in the training distribution, shows that the number of collisions is also somewhat reduced within the training distribution. We hypothesize that the reason for this effect is that some situations that cause collisions are seldom seen during the training process, and therefore the epistemic uncertainty in those situations remains high.

\begin{figure}[!t]
    \centering
    \subfloat[Collisions, fixed $K=10$.]
    {\includegraphics[width=0.48\columnwidth]{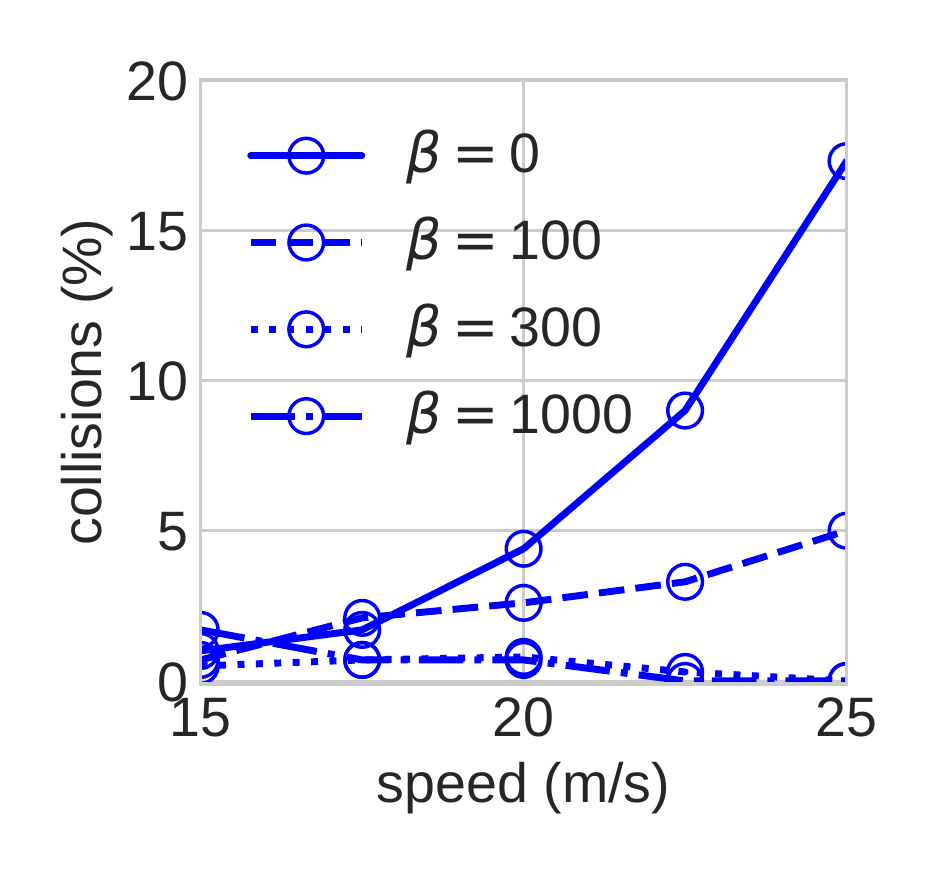}
    \label{fig:high_speed_cols_var_beta}}
    \hfil
    \subfloat[Collisions, fixed $\beta=300$.]
    {\includegraphics[width=0.48\columnwidth]{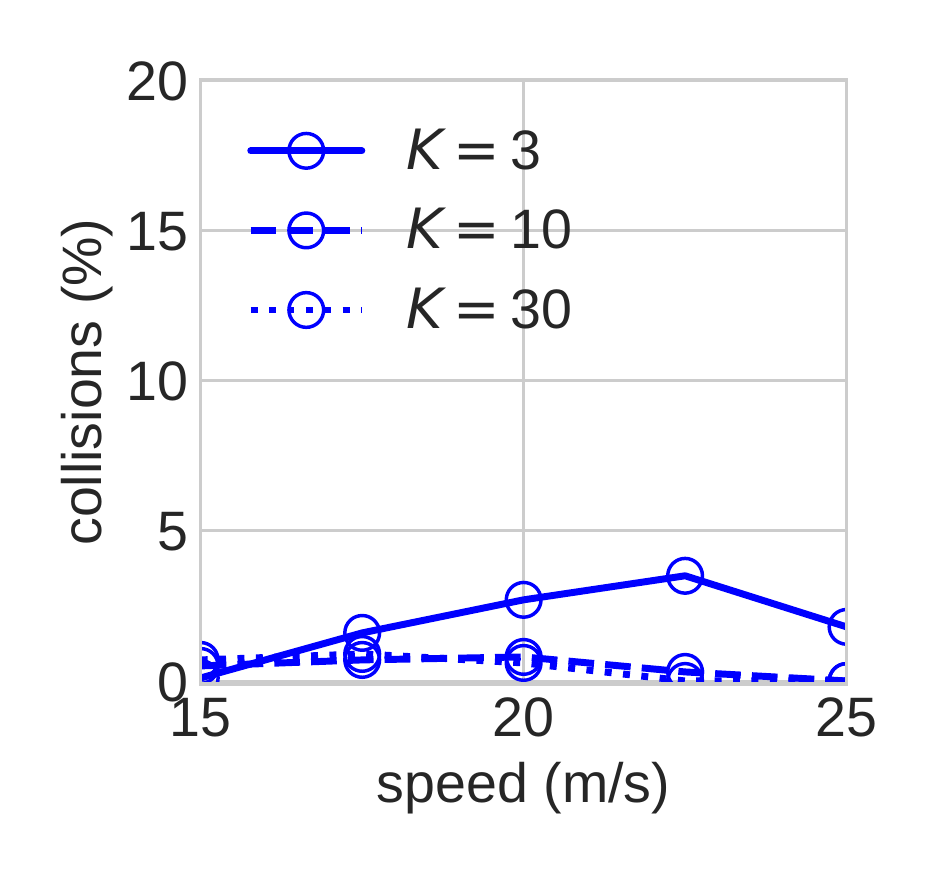}
    \label{fig:high_speed_cols_var_K}}
    \hfil
    \subfloat[Timeouts, fixed $K=10$.]
    {\includegraphics[width=0.48\columnwidth]{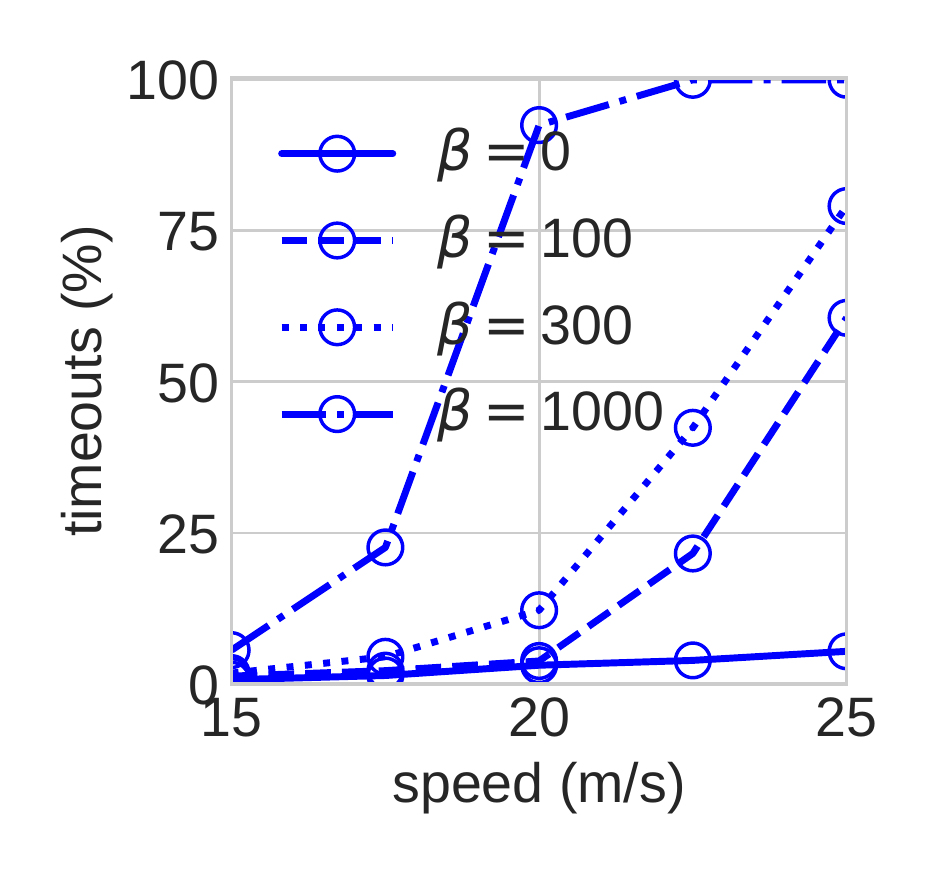}
    \label{fig:high_speed_timeouts_var_beta}}
    \hfil
    \subfloat[Timeouts, fixed $\beta=300$.]
    {\includegraphics[width=0.48\columnwidth]{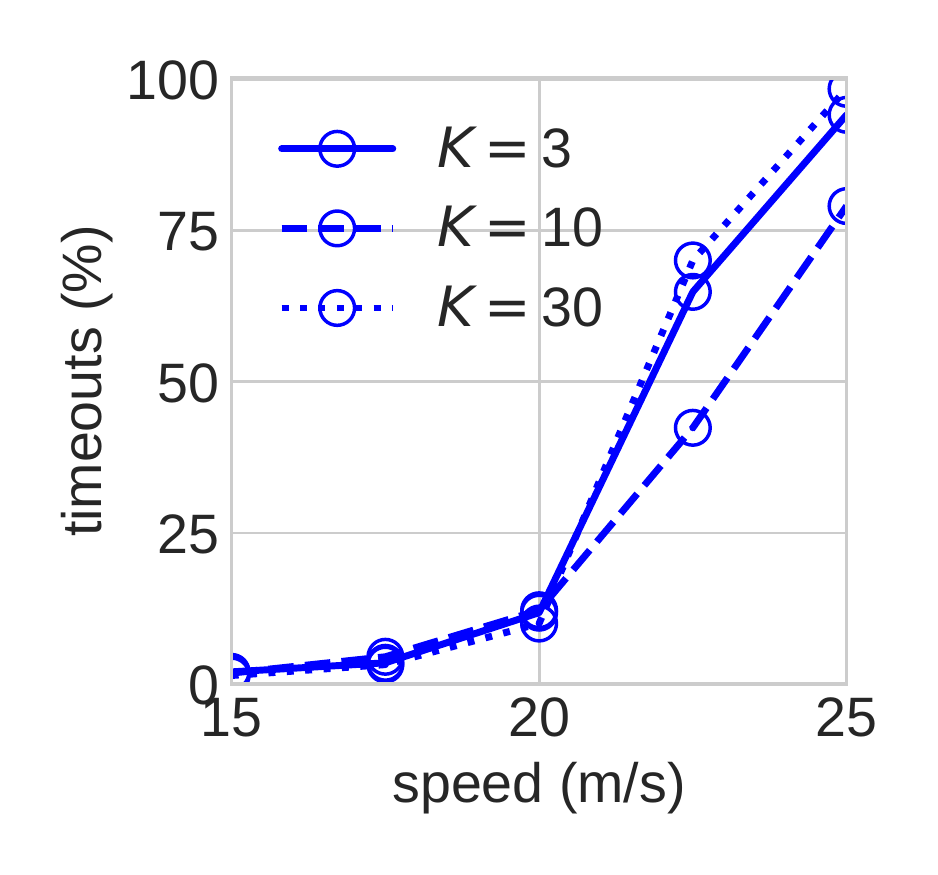}
    \label{fig:high_speed_timeouts_var_K}}
    \caption{Number of collisions and timeouts for the RPF agent, with uncertainty threshold $\sigma_\mathrm{e}=2$, varying values of the prior scaling factor $\beta$ and the number of ensemble members $K$, in situations outside of the training distribution.}
    \label{fig:increased_speed_parameters}
\end{figure}

As previously mentioned, the scaling factor $\beta$ and the number of networks $K$ do not substantially influence the performance of the RPF agent inside the training distribution. However, these parameters determine how well the agent can estimate the epistemic uncertainty and detect dangerous situations, which is illustrated in Fig.~\ref{fig:increased_speed_parameters}. In short, $\beta$ and $K$ need to be sufficiently large to give a reasonable uncertainty estimate. How to set these parameter values are further discussed in Sect.~\ref{sec:discussion}.

\subsection{Aleatoric and epistemic uncertainty estimation}

\begin{figure}[!t]
    \centering
    \includegraphics[width=0.98\columnwidth]{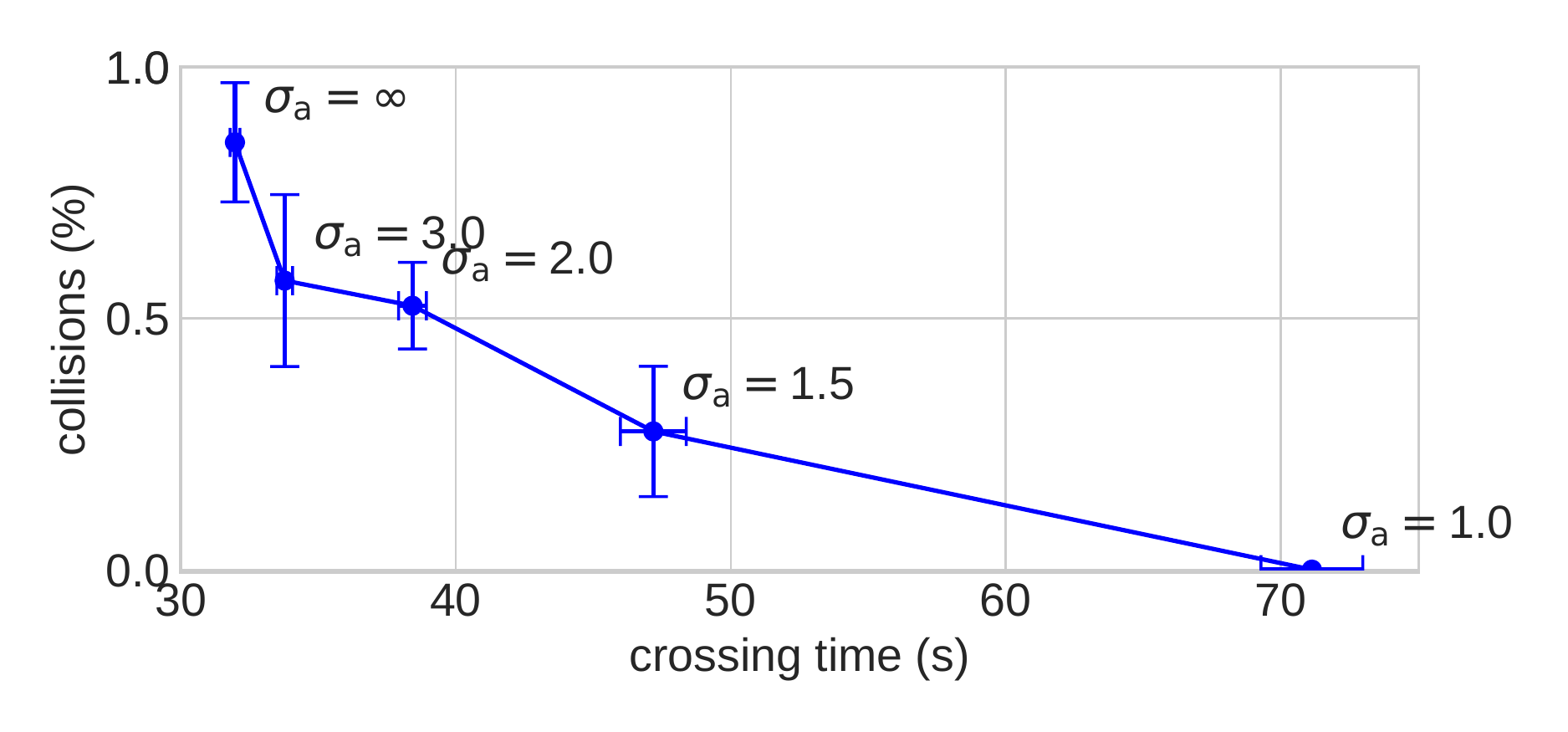}
    \caption{Number of collisions and crossing time for the EQN algorithm for different levels of allowed aleatoric uncertainty, which is achieved by varying the parameter $\sigma_\mathrm{a}$.}
    \label{fig:iqnrpf_pareto_sigma_a}
\end{figure}

The EQN agent performs better than the RPF agent and similar to the IQN agent within the training distribution, see Table~\ref{tab:main_scenario}. Importantly, the EQN agent combines the advantages of the other two agents and can estimate both the aleatoric and epistemic uncertainty of a decision. When the aleatoric uncertainty criterion is applied, the number of situations that are classified as uncertain depends on the parameter $\sigma_\mathrm{a}$, see Fig.~\ref{fig:iqnrpf_pareto_sigma_a}. Thereby, the trade-off between risk and time efficiency, here illustrated by number of collisions and crossing time, can be controlled by tuning the value of $\sigma_\mathrm{a}$.

\begin{figure}[!t]
    \centering
    \subfloat[Collisions]
    {\includegraphics[width=0.98\columnwidth]{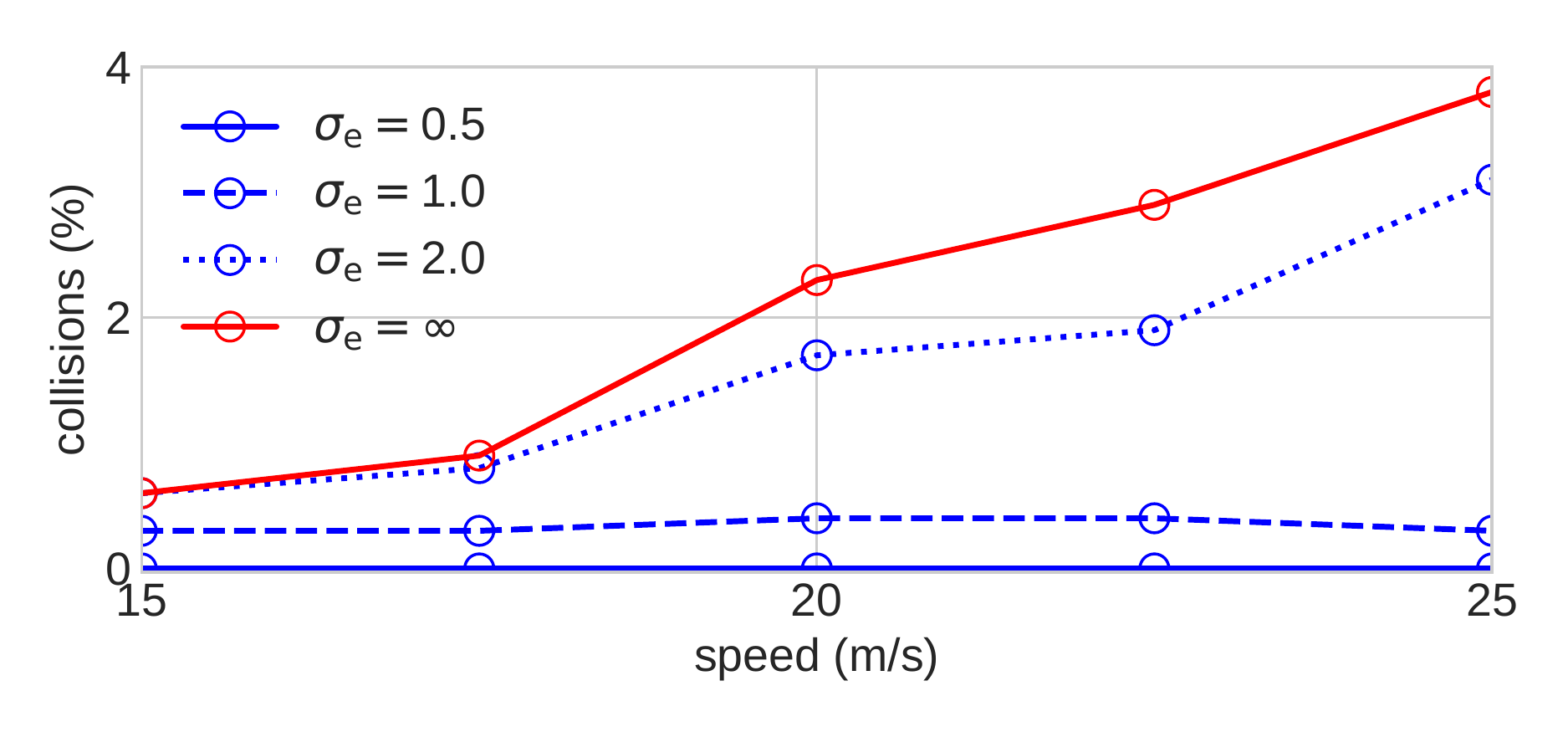}}
    \hfil
    \subfloat[Timeouts]
    {\includegraphics[width=0.98\columnwidth]{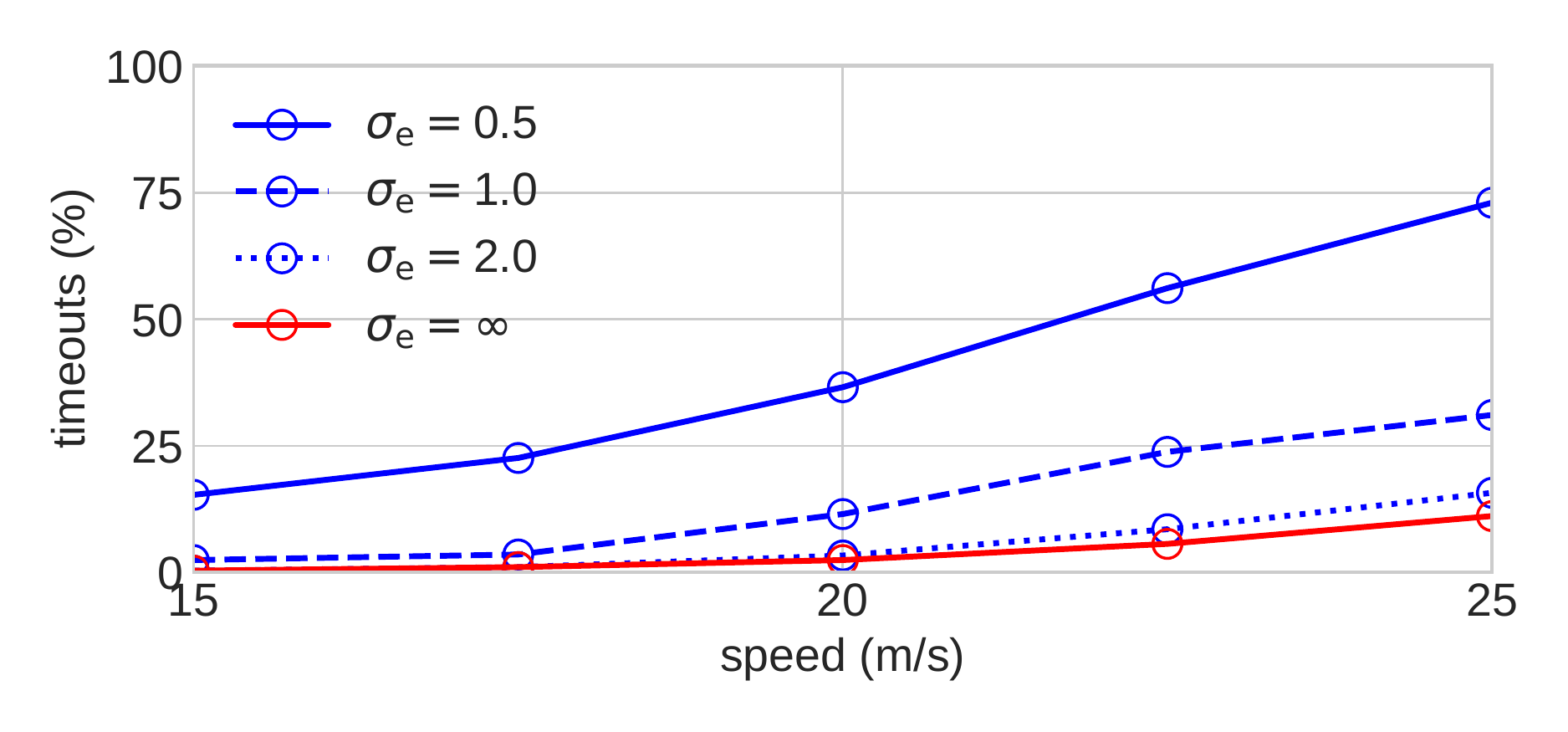}}
    \caption{Number of collisions and timeouts for the EQN agent in situations outside the training distribution. The maximum speed of the crossing vehicles is $15$ m/s during the training process, and then the speed is gradually increased in the testing episodes.}
    \label{fig:iqnrpf_increased_speed}
\end{figure}

The performance of the epistemic uncertainty estimation of the EQN agent is illustrated in Fig.~\ref{fig:iqnrpf_increased_speed}, where the speed of the surrounding vehicles is increased. Similarly as for the RPF agent, a sufficiently strict epistemic uncertainty criterion, i.e., sufficiently low value of the parameter $\sigma_\mathrm{e}$, prevents the number of collisions to increase when the speed of the surrounding vehicles increases. The result at $15$ m/s also indicates that the number of collisions within the training distribution is somewhat reduced when the epistemic uncertainty condition is applied.
Interestingly, when combining moderate aleatoric and epistemic uncertainty criteria, by setting $\sigma_\mathrm{a}=1.5$ and $\sigma_\mathrm{e}=1.0$, all the collisions within the training distribution are removed, see Table~\ref{tab:main_scenario}. These results show that it is useful to consider the epistemic uncertainty even within the training distribution, where the detection of uncertain situations can prevent collisions in rare edge cases.

\section{Discussion}
\label{sec:discussion}

The results show that the IQN and RPF agents can provide estimates of the aleatoric and epistemic uncertainties, respectively. When combined with the uncertainty criteria, situations with high uncertainty are identified, which can be used to make safer decisions. Further use and characteristics of the uncertainty information are discussed below. The results also demonstrate that the EQN agent combines the advantages of the individual components and provides a full uncertainty estimate, including both the aleatoric and epistemic dimensions.

\begin{figure}[!t]
    \centering
    \includegraphics[width=0.8\columnwidth]{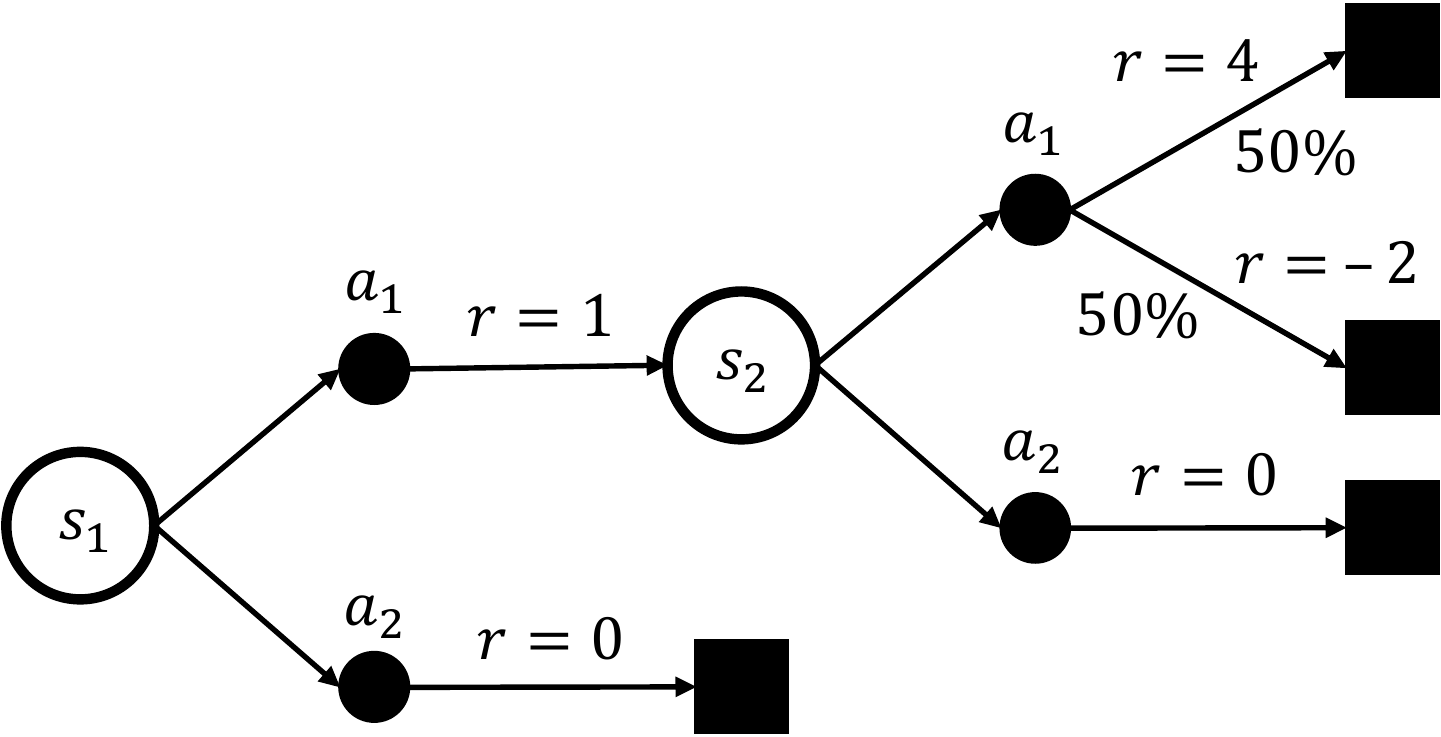}
    \caption{A simple MDP, which illustrates that a risk-sensitive IQN policy needs to be retrained for each value of the CVaR parameter $\alpha$.}
    \label{fig:simple_mdp}
\end{figure}

The aleatoric uncertainty estimate, given by the IQN or EQN algorithms, can be used to balance risk and time efficiency, either by training in a risk-sensitive way (varying the CVaR parameter $\alpha$, see Fig.~\ref{fig:pareto_alpha}) or applying the aleatoric uncertainty criterion (varying the allowed variance $\sigma_\mathrm{a}^2$, see Fig.~\ref{fig:pareto_sigma_a} and~\ref{fig:iqnrpf_pareto_sigma_a}). An important advantage of the uncertainty criterion approach is that its parameter $\sigma_\mathrm{a}$ can be tuned after the training process has been completed, whereas the agent needs to be retrained for each CVaR parameter $\alpha$. However, the uncertainty criterion only works in practical applications, such as autonomous driving, where a backup policy can be defined. The CVaR approach does not require a backup policy and is therefore suitable for environments where such a policy is hard to define, e.g., the \mbox{Atari-57} benchmark~\cite{Dabney2018}.
Bernhard et al. first trained a risk-neutral IQN agent, then lowered the CVaR threshold after the training process had been completed, and showed a reduction of collisions in an intersection driving scenario~\cite{Bernhard2019}. However, such a procedure does not provide the correct estimate of the return distribution $Z_\tau(s,a)$ for a risk-averse setting ($\alpha < 1$) and could lead to arbitrary decisions. The problem with this approach is easily seen for the simple MDP shown in Fig.~\ref{fig:simple_mdp}. The risk-neutral policy is $\pi_\mathrm{rn}(s_1)=a_1$ and $\pi_\mathrm{rn}(s_2)=a_1$, whereas the risk-averse policy is $\pi_\mathrm{ra}(s_1)=a_1$ and $\pi_\mathrm{ra}(s_2)=a_2$, for CVaR parameter $\alpha < 0.75$. For this risk-averse policy, the return of the initial state is $Z^{\pi_\mathrm{ra}}(s_1, \pi_\mathrm{ra}(s_1))=1$, with probability $1$. However, if $Z_\tau(s, a)$ is first estimated for the risk-neutral policy, and then the action that maximizes the CVaR for $\alpha < 0.6$ is chosen, this risk-averse policy gives $\tilde{\pi}_\mathrm{ra}(s_1)=a_2$, and the return of the initial state is $Z^{\tilde{\pi}_\mathrm{ra}}(s_1, \tilde{\pi}_\mathrm{ra}(s_1))=0$, with probability $1$. In short, the policy $\tilde{\pi}_\mathrm{ra}$ becomes sub-optimal, since it considers the aleatoric uncertainty of the risk-neutral policy, which is irrelevant in this case.

An alternative to estimating the distribution over returns and still consider aleatoric risk in the decision-making is to adapt the reward function. Risk-sensitivity could be achieved by, for example, increasing the size of the negative reward for collisions. However, rewards with different orders of magnitude create numerical problems, which can disrupt the training process~\cite{Mnih2015}.
Furthermore, for a complex reward function, it would be non-trivial to balance the different components to achieve the desired result. 

The epistemic uncertainty information provides insight into how far a situation is from the training distribution. In this study, the usefulness of an epistemic uncertainty estimate is demonstrated by increasing the safety, through classifying the agent's decisions in situations far from the training distribution as unsafe and then instead applying a backup policy. If it is possible to formally guarantee safety with a learning-based method is an open question, and likely an underlying safety layer is required in a real-world application~\cite{Underwood2016}. The RPF and EQN agents can reduce the activation frequency of such a safety layer, but possibly even more importantly, the epistemic uncertainty information could be used to guide the training process to regions of the state space in which the current agent requires more training. Furthermore, if an agent is trained in a simulated world and then deployed in the real world, the epistemic uncertainty information can identify situations with high uncertainty, which should be added to the simulated world.

The algorithms that were introduced in this paper include a few hyperparameters, whose values need to be set appropriately. The aleatoric and epistemic uncertainty criteria parameters, $\sigma_\mathrm{a}$ and $\sigma_\mathrm{e}$, can both be tuned after the training is completed and allow a trade-off between risk and time efficiency, see Fig.~\ref{fig:pareto_sigma_a}, \ref{fig:increased_speed}, \ref{fig:iqnrpf_pareto_sigma_a}, and~\ref{fig:iqnrpf_increased_speed}. Note that both these parameters determine the allowed spread in returns, between quantiles or ensemble members, which means that the size of these parameters are closely connected to the magnitude of the reward function.
In order to detect situations with high epistemic uncertainty, a sufficiently large spread between the ensemble members is required, which is controlled by the scaling factor $\beta$ and the number of ensemble members $K$. The choice of $\beta$ scales with the magnitude of the reward function. A too small parameter value creates a small spread, which makes it difficult to classify situations outside of the training distribution as uncertain, see Fig.~\ref{fig:high_speed_cols_var_beta} and~\ref{fig:high_speed_timeouts_var_beta}. On the other hand, a too large value of $\beta$ makes it difficult for the trainable network to adapt to the fixed prior network.
Furthermore, an increased number of ensemble members $K$ naturally increases the accuracy of the epistemic uncertainty estimate, see Fig.~\ref{fig:high_speed_cols_var_K} and~\ref{fig:high_speed_timeouts_var_K}, but induces a higher computational cost.

All the tested methods have a similar sample complexity, but the uncertainty-aware approaches require more computational resources than the baseline DQN method. The IQN method uses $N$ quantile samples in the loss function, the RPF method trains an ensemble of $K$ neural network, and the EQN method combines these two features. However, the design of the algorithms allows a parallel implementation, which in practice reduces the difference in training time. All agents were trained on a standard desktop computer, where the DQN agent required $12$ hours, the IQN agent $24$ hours, the RPF agent $72$ hours, and the EQN agent $96$ hours. However, since the focus of this study is not to optimize the implementation, the time efficiency can be significantly improved.

\section{Conclusion}
\label{sec:conclusion}

The results show that the proposed EQN algorithm combines the advantages of the IQN and RPF methods, and can thereby provide a complete uncertainty estimate of its decisions, including both the aleatoric and the epistemic uncertainty. The aleatoric uncertainty criterion allows an agent to balance risk and time efficiency after the training is completed and achieves similar results as an agent that is trained in a risk-sensitive way, with the benefit that the agent does not need to be retrained for each uncertainty threshold. Furthermore, the results show that the epistemic uncertainty criterion can be used to identify situations that are far from the training distribution, in which the agent could make dangerous decisions. The awareness of such situations can be used to enhance the safety of the trained agent and to improve the training process.

The EQN algorithm provides a general approach to create an uncertainty-aware decision-making agent for autonomous driving. However, in order to apply the method to other driving scenarios than the intersections that were considered in this study, the MDP formulation needs to be adapted to the new scenarios, or an MDP that covers multiple scenarios needs to be constructed. While the DQN-family of methods have proved to work well for different types of driving scenarios~\cite{Zhu2021, Ye2021}, future work involves to test the EQN method in more scenarios and other simulation environments, before performing tests in the real world. It would also be interesting to investigate how the algorithm would handle different aspects of noise in the sensor signals.
Another topic of future work is to investigate how the epistemic uncertainty estimation can be used to bridge the gap between simulators and reality. An RPF or EQN agent that has been trained in a simulated world could potentially detect traffic situations in the real world where the epistemic uncertainty is high and then automatically add these situations to the simulated environment.

\ifCLASSOPTIONcaptionsoff
  \newpage
\fi

\linespread{0.98}
\bibliography{references}{}
\bibliographystyle{ieeetr}

\vspace{-25pt}

\begin{IEEEbiography}[{\includegraphics[width=1in,height=1.25in,clip,keepaspectratio]{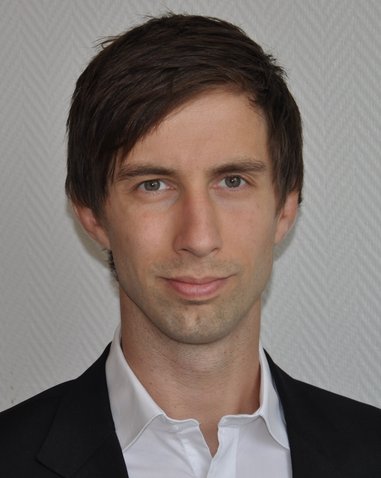}}]{Carl-Johan Hoel}
received the B.S. and M.S. degrees in physics from Chalmers University of Technology, Gothenburg, Sweden. He is currently working towards the Ph.D. degree at Chalmers University of Technology, together with the Volvo Group, Gothenburg, Sweden. His research focuses on robust reinforcement learning methods for creating general decision-making agents, applied to autonomous driving.
\end{IEEEbiography}

\vspace{-25pt}

\begin{IEEEbiography}[{\includegraphics[width=1in,height=1.25in,clip]{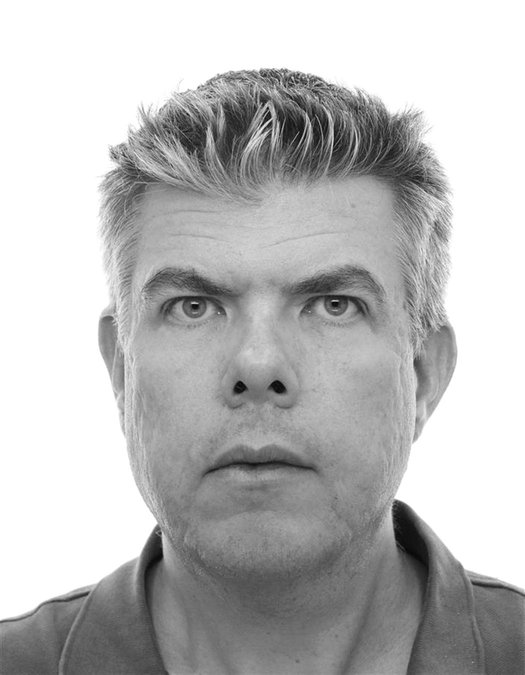}}]{Krister Wolff}
received the M.S. degree in physics from Gothenburg University, Gothenburg, Sweden and the Ph.D. degree from Chalmers University of Technology, Gothenburg, Sweden. He is currently an Associate Professor of adaptive systems, and he is also the Vice head of Department at Mechanics and maritime sciences, Chalmers. His research is within the application of AI in different domains, such as autonomous robots and self-driving vehicles, using machine learning and bio-inspired computational methods as the main approaches. 
\end{IEEEbiography}

\vspace{-25pt}

\begin{IEEEbiography}[{\includegraphics[width=1in,height=1.25in,clip]{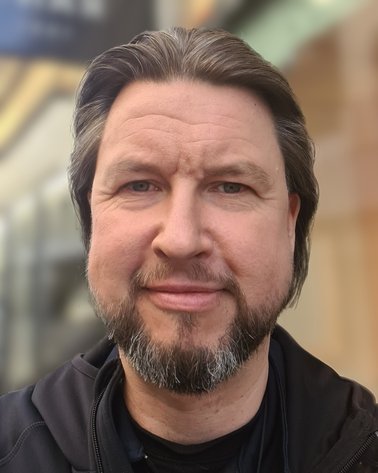}}]{Leo Laine}
received the Ph.D. degree from Chalmers University of Technology, Gothenburg, Sweden, within Vehicle Motion management. Since 2007, he has been with the Volvo Group Trucks Technology (VGTT) in the Vehicle Automation department. Since 2013, he has also been an Adjunct Professor in vehicle dynamics with Chalmers Vehicle Engineering and Autonomous Systems. Since 2013, he is a specialist within complete vehicle control. Since 2019, he is a technical advisor within Vehicle Motion and Energy Management within VGTT.

\vspace{-25pt}

\end{IEEEbiography}

\end{document}